\documentclass[10pt,a4paper,twocolumn]{article}

\usepackage[T1]{fontenc}
\usepackage[utf8]{inputenc}
\usepackage{lmodern}
\usepackage[british]{babel}
\usepackage{microtype}

\usepackage[
  top=19mm,
  bottom=20mm,
  left=22mm,
  right=18mm
]{geometry}
\usepackage{titlesec}
\titleformat{\section}{\large\bfseries\raggedright}{\thesection}{0.75em}{}
\titleformat{\subsection}{\normalsize\bfseries\raggedright}{\thesubsection}{0.75em}{}
\usepackage{balance}
\usepackage{amsmath,amssymb}
\usepackage{graphicx}
\graphicspath{{figures/}}
\usepackage{booktabs,tabularx,longtable}
\usepackage{array}
\usepackage{placeins}
\usepackage[font=small,labelfont=bf]{caption}

\usepackage{csquotes}
\usepackage[
  backend=biber,
  style=authoryear,
  giveninits=true,
  uniquename=init,
  maxcitenames=2,
  maxbibnames=99,
  doi=true,
  isbn=false,
  url=true
]{biblatex}
\DeclareLabeldate[online]{%
  \field{date}
  \field{year}
  \field{eventdate}
  \field{origdate}
  \literal{nodate}
}
\AtBeginBibliography{\small\setlength{\bibitemsep}{3pt}\setlength{\bibparsep}{0pt}}
\usepackage{xurl}
\usepackage[pdfusetitle,hidelinks]{hyperref}
\hypersetup{pdftitle={APEX: An Extensible Model for Agent-Assisted Production Scheduling},pdfsubject={An extensible production model, hybrid multiobjective search, and agent-assisted scheduling and model refinement},pdfkeywords={production scheduling, extensible models, model refinement, job shop, flexible job shop, permutation flow shop, multiobjective optimisation, hybrid metaheuristics, agent-assisted planning, benchmarking}}
\newcommand{\MS}{\operatorname{MS}}
\newcommand{\FT}{\operatorname{FT}}
\newcommand{\HV}{\operatorname{HV}}

\begin{document}

% Keep the title and abstract together above the two-column main text.
\twocolumn[\begin{@twocolumnfalse}
\title{APEX: An Extensible Model for\\
Agent-Assisted Production Scheduling}
\author{Felix J. Grumbach\qquad Stefan Görlitz\\[0.45em]
\normalsize Qunevo GmbH, Germany\\[0.25em]
\small \href{mailto:research@qunevo.com}{\nolinkurl{research@qunevo.com}}}
\hypersetup{pdfauthor={Felix J. Grumbach and Stefan Görlitz}}
\date{26 September 2026}
\maketitle

\begin{abstract}
Production scheduling requires realistic models that reflect operational constraints
and efficient methods that balance competing goals. Putting these methods into use
also requires data integration, model adaptation and specialist expertise.
We present APEX, an extensible production scheduling framework built around
a general model and hybrid multiobjective search. Agent assistance supports
both scheduling and model refinement: agents prepare data and explore
scenarios in natural language, while coding agents help implement and test
new constraints and objectives. Shared construction and checking procedures
connect these adaptations to the scheduling core.
We benchmark eight APEX configurations against NSGA-II, SPEA2, MOEA/D and
SMS-EMOA on 69 public job-shop, flexible job-shop and permutation flow-shop
instances, assessing workload completion time (makespan), total job flowtime
and computation time.
Hybrid configurations achieve the best aggregate solution quality, although
the leading method depends on the problem class and objective.
A separate synthetic workflow study uses OpenAI's GPT-6-astra as an
interaction layer between the human planner and the algorithmic core,
testing rule additions, plan and objective changes, and what-if comparisons.
All 24 sessions
completed the requested changes and passed independent checks of saved
models and schedules. A separate coding evaluation produced six native
implementations of an additional objective or hard constraint through
predefined extension hooks. All passed independent checks without modifying
the core.
\end{abstract}

\par\vspace{0.5em}
{\small\noindent\textbf{Acknowledgements.}
The authors acknowledge the foundational research in
\href{https://its-owl.de/projekte/support/}{SUPPORT} (project ref.\ \mbox{005-2111-0026}) and the
\href{https://www.hsbi.de/forschung/forschungsprojekte/aktuelle-projekte-fb-3/schenck-smart-service-lab}{Human-Centred Smart Service Lab} (project ref.\ \mbox{EFRE-0300180}),
particularly \href{https://www.hsbi.de/forschung/forschungsprojekte/aktuelle-projekte-fb-3/reusch-predictive-scheduling}{Predictive Scheduling},
which provided valuable insights for the development of APEX.
They thank the partners involved in this earlier research: Hochschule Bielefeld,
Fraunhofer IOSB-INA, Miele, Isringhausen, Bio-Circle Surface Technology,
MIT Moderne Industrietechnik and PerFact Innovation.
They also acknowledge the
\href{https://its-owl.de/news-events/news/qunevo-start-up-aus-owl-sorgt-mit-ki-gestuetzter-produktionsplanung-fuer-mehr-effizienz-und-flexibilitaet/}{Qunevo spin-off project}
(project ref.\ EFRE~20800400) for supporting the transfer of research into practice.\par}

\vspace{1.5em}
\end{@twocolumnfalse}]

\section{Introduction}\label{sec:introduction}

Production planners repeatedly solve coupled combinatorial problems:
sequencing work, assigning scarce resources and deciding when activities
can start. Jobs compete for machines, people and materials, while breakdowns,
late deliveries and urgent orders can invalidate earlier decisions. Planners
must reconcile these changes with work already released. When the model or
its information no longer matches the shop floor, repeated manual
intervention becomes routine firefighting
\parencite[pp.~1--4]{Grumbach2024}.

A usable schedule requires both an adequate model and current information.
A machine may be free while its preparation technician is unavailable, or
an operation may finish before inspection releases its output.
\textcite{Eriksson2022} illustrate the operational gap in a manufacturing
case: daily updates of an advanced planning and scheduling (APS) system
lagged intraday changes, and employees repeatedly overrode its plans.
Trust and organisational difficulties also affected its use.

Adapting APS systems to production can be costly, requiring data integration,
model customisation and continuing specialist support. In three manufacturing
cases involving tactical APS planning, \textcite{KjellsdotterIvert2011}
document data and model-design problems, consultant dependence and substantial
operating effort, including vendor-assisted algorithm customisation.
Specialised mathematical models and heuristics also need experts to identify
relevant constraints and objectives, design suitable search procedures and
check operational validity. An available optimisation engine therefore
addresses only part of the implementation task.

Agent-assisted modelling offers a way to support this work. OptiMUS uses
language-model agents to formulate optimisation models and generate solver
code \parencite{Ahmaditeshnizi2024}. Detailed production scheduling additionally
needs a reusable model and execution framework that preserve constraints
across planning cycles, scenario changes and search methods. Such a framework
can give agents a common basis for operating the planner and adapting its
capabilities as production requirements evolve.

APEX builds on several years of research and industrial collaboration,
including flexible machine and worker scheduling \parencite{Grumbach2023}
and human-centred assembly sequencing \parencite{Vollenkemper2023}.
\textcite{Grumbach2024} shows how generic, data-driven simulation represents
richer production constraints and how machine learning supports scheduling
under uncertainty. Integrating these methods into a field-synchronous
overall system remains an open research and development task: a digital
planning twin that aligns its model with the shop floor and connects
planning decisions with execution feedback.

APEX develops this line of work through a general production model that can
be configured and extended as requirements evolve. Workplans, resources,
calendars, materials and commitments provide a reusable representation
across production settings. Agent assistance spans both scheduling and
model refinement: agents prepare data, compare scenarios and invoke search;
coding agents help implement and test requirements beyond the existing
capabilities. This refinement ranges from configuring supported rules to
adding new constraints and objectives. Shared construction, evaluation and validation
procedures connect these changes to the complementary search methods,
allowing the model to evolve with the application.

The paper examines two connected aspects of this framework. The computational
study asks: \emph{How do APEX's search components and their combinations trade
solution quality against computation time compared with established
multiobjective algorithms?} We compare eight APEX configurations with four
external algorithms on 69 public job-shop (JSP), flexible job-shop (FJSP) and
permutation flow-shop (PFSP) instances
\parencite{Brandimarte1993,Beasley1990,Taillard1993},
using makespan and total job flowtime. A separate synthetic study asks:
\emph{Can an agent carry out requested rule, plan and objective changes,
and compare what-if scenarios, while preserving the intended semantics?}
Eight specified requests, each repeated in three fresh contexts, are checked
against the saved models, schedules and action traces. This development
study examines supported configuration and planning operations. A second
experiment asks whether a coding agent can implement an additional objective
and a new hard constraint through the customization interface, with independent
acceptance checks of the submitted implementations.

Section~\ref{sec:related-work} reviews production models, solution methods and
agent assistance. Sections~\ref{sec:theory} and~\ref{sec:search} describe the
model and search components; Sections~\ref{sec:data-methods} and~\ref{sec:results}
present the algorithm benchmark. Section~\ref{sec:discussion} examines the
four agent-assisted workflows and the native-extension experiment, then relates
them to a proposed digital planning twin, followed by limitations and
conclusions.

\section{Production scheduling models, solution methods and agent assistance}\label{sec:related-work}

Assessing a scheduling system requires three separate questions: which
production conditions its model represents, how it searches for good schedules,
and how it connects those schedules to operational decisions. This section
reviews selected work on all three, including recent contributions from
2025--2026, to position APEX's model, search and agent interface.

\subsection{Production models and operational realism}

Production forms impose different scheduling structures. Flow shops follow a
common sequence of stages; hybrid flow shops provide parallel machines within
stages. Job shops allow job-specific operation sequences, and flexible job shops
add alternative processing machines. Assembly and project scheduling introduce
more general precedence relations and shared resource capacities. A common
representation can span several such families: the PyJobShop preprint by
\textcite{Lan2025} uses jobs, tasks, execution modes, resources and temporal
relations to connect machine and project scheduling through one modelling
interface.

Recent models also address requirements beyond machine assignment and operation
sequencing. \textcite{Kasapidis2025} combine renewable and nonrenewable resources,
finite buffers and blocking within a unified flexible-job-shop framework.
\textcite{Perrachon2025} divide operations into stages so that a resource is
required for only part of an operation. This captures, for example, an operator
who starts a machine and returns after unattended processing. These are direct
precedents for generic models with interacting production requirements.

Table~\ref{tab:scheduling-models} illustrates further extensions, including
operator workload, energy, transport and uncertainty. The examples distinguish
the represented conditions from the setting in which a method was evaluated.
The distinction matters even for apparently familiar additions:
\textcite{Terbrack2025} find that including setups and due-date compliance
restricts attainable energy improvements. The interaction between requirements
can therefore change the conclusions obtained from a simpler model.

\begin{table*}[!t]
\centering\small
\caption{Selected scheduling models and solution approaches. CP denotes
constraint programming; MILP denotes mixed-integer linear programming.}
\label{tab:scheduling-models}
\begin{tabularx}{\textwidth}{@{}>{\raggedright\arraybackslash}p{.21\textwidth}>{\raggedright\arraybackslash}p{.38\textwidth}>{\raggedright\arraybackslash}X@{}}
\toprule
Study and setting & Model coverage & Method and evidence setting \\
\midrule
\textcite{Lan2025}; multiple machine and project families & Common task, mode, resource and temporal-constraint interface & CP; cross-family public benchmarks; preprint \\[.45em]
\textcite{Kasapidis2025}; flexible job shop & Tools, utilities, material resources, buffers and blocking & CP with adaptive large-neighbourhood search; established and generated combined-resource instances \\[.45em]
\textcite{Perrachon2025}; multi-resource job shop & Resources required during selected operation stages & Simulated annealing with reinsertion; benchmark comparison with commercial CP \\[.45em]
\textcite{Yasari2025}; operator--machine job shop & Operator assignment and physical-workload limits & MILP with row generation; extended benchmark instances \\[.45em]
\textcite{Terbrack2025}; flexible job shop & Energy costs, demand peaks and emissions, with setups and due dates & CP with lexicographic objectives; computational instances \\[.45em]
\textcite{Ferreira2026}; hybrid flow shop & Parallel machines at stages; sequence- and machine-dependent setups & Column generation and iterated local search; generated two-stage instances \\[.45em]
\textcite{Chen2025}; dynamic flexible job shop & New arrivals, single-crane transport and sequence-dependent setups & Evolutionary multitask search and genetic programming; abstract-level method account \\[.45em]
\textcite{Zhang2026}; uncertain job shop & Job routes branch according to uncertain events & Actor--critic reinforcement learning; generated stochastic instances \\
\bottomrule
\end{tabularx}
\end{table*}

\subsection{Solution methods and computational trade-offs}

Mathematical programming and CP remain relevant alongside specialised
heuristics. They express feasibility explicitly and can provide bounds and,
when search completes, optimality certificates. Their practical performance
depends on the formulation, problem family and available computation. In the
PyJobShop experiments, solver rankings vary across families and sizes, and
formulation choices materially affect performance \parencite{Lan2025}.
A method name alone is therefore an insufficient basis for choosing a solver.

Scale provides another dimension of realism. \textcite{DaCol2022} test
constraint-programming solvers on generated job shops with up to one million
operations, using fixed routes, precedence and machine capacities. Scale and
constraint coverage address different dimensions of a production model.

Hybrid methods combine this modelling machinery with targeted search.
\textcite{Kasapidis2025} use adaptive large-neighbourhood search to guide CP
through selected subproblems. For hybrid flow shops, \textcite{Ferreira2026}
combine column generation with an iterated local search that supplies initial
columns and an upper bound. Simulated annealing provides another route:
\textcite{Perrachon2025} adapt reinsertion moves and feasibility filtering to
their richer resource model. These examples show why model design and search
design must be considered together: useful moves, decompositions and checks
depend on the constraints being represented.

Learning can also prepare decisions for subsequent scheduling without using
language models. \textcite{Chen2025} combine evolutionary multitask optimisation
with genetic programming, which constructs scheduling rules for a dynamic
environment. \textcite{Zhang2026} train an actor--critic policy for jobs whose
future routes depend on uncertain events. Such branching represents uncertainty
about what processing will be needed, rather than a routing alternative chosen
by the planner. For learned rules and policies, evaluation should distinguish
training or search effort from later decision cost and test performance when
the operating conditions change. For any method, a useful comparison must
match the model, objectives, information available at decision time and
computational deadline.

\subsection{Agent-assisted modelling and operation}

Agent assistance adds capabilities around these models and methods, and can
also participate in heuristic design. An agent need not use a large language
model (LLM): \textcite{Bakopoulos2024} coordinate heuristic,
mathematical-programming and reinforcement-learning schedulers with a
production-manager interface and digital-twin validation. LLM-based OptiMUS
instead addresses formulation work: specialised agents translate descriptions
into linear or mixed-integer linear models, generate and debug solver code,
and evaluate executions \parencite{Ahmaditeshnizi2024}.

Other agents operate existing tools or develop new rules. MASC selects
scheduling algorithms and coordinates observation, scheduling, planning and
control \parencite{Wang2025}. The actor--critic approach of
\textcite{May2026} uses LLMs to generate dispatching rules and critique them
with simulation feedback. This differs from numerical reinforcement learning
such as \textcite{Zhang2026}, where the trained policy itself selects actions.

At the interaction level, \emph{Chat with MES} interprets requests and plans
operations on a simulated manufacturing execution system \parencite{Yuan2025}.
The preprint by \textcite{Ye2026} connects natural-language preferences,
genetic search and explanations of Pareto alternatives; its architecture is
used here at abstract level. Agents can also explore simulation parameters
\parencite{Xia2024} or orchestrate optimisation and discrete-event simulation
for planning scenarios \parencite{Korth2026}. Table~\ref{tab:agent-studies}
summarises the different evidence settings. Success in operating a tool,
improving a simulated schedule and supporting an industrial planner are
distinct outcomes.

\begin{table*}[htbp]
\centering\small
\caption{Selected agent-assisted approaches. The studies address different
tasks and do not constitute a common performance comparison.}
\label{tab:agent-studies}
\begin{tabularx}{\textwidth}{@{}>{\raggedright\arraybackslash}p{.25\textwidth}>{\raggedright\arraybackslash}p{.28\textwidth}>{\raggedright\arraybackslash}X@{}}
\toprule
Study & Agent role & Evidence setting \\
\midrule
\textcite{Bakopoulos2024} & Coordination of scheduling methods & Bicycle production; digital-twin validation \\[.4em]
\textcite{Ahmaditeshnizi2024} & Model formulation and solver-code generation & Natural-language LP/MILP modelling benchmarks \\[.4em]
\textcite{May2026} & Generation and critique of dispatching rules & Dynamic job-shop simulation \\[.4em]
\textcite{Wang2025} & Scheduling-tool selection and coordination & FJSP simulations and robotic experiments \\[.4em]
\textcite{Yuan2025} & Natural-language system operation & Simulated garment MES; designed requests \\[.4em]
\textcite{Ye2026} & Preference interpretation and Pareto explanation & Job shops, flow shops and flexible job shops; preprint \\[.4em]
\textcite{Xia2024} & Simulation parameter exploration & Digital-twin proof of concept \\[.4em]
\textcite{Korth2026} & Orchestration of optimisation and simulation & Supplier-data planning scenarios \\
\bottomrule
\end{tabularx}
\end{table*}

\subsection{Research gap and positioning of APEX}

This literature makes the concern raised by \textcite{Grumbach2024} more
specific: improving an algorithm within a fixed abstraction leaves open
whether that abstraction captures the constraints that determine feasibility
and usefulness on the shop floor. Rich generic models already exist, including
models that combine several resource restrictions. The remaining challenge is
to represent the relevant conditions jointly, keep their data and assumptions
aligned with production, and evaluate the resulting decisions in operation.
The APS experience reported by \textcite{Eriksson2022} illustrates why timely
state updates, manual intervention and organisational use matter alongside
the scheduling algorithm.

For a field-synchronous digital planning twin, this implies a continuing link
between observed production state, the planning model, decisions and execution
feedback. Model coverage, solution quality under a computational deadline,
and operational validation should therefore be assessed separately. APEX
addresses this agenda through a shared, extensible production model and search
components that construct and improve schedules under common checking
semantics. Agent assistance supports data preparation, scenario work and model
extension around this core; XH, XT and XE themselves perform numerical search.

The paper presents the richer model, benchmarks its search on classical JSP,
FJSP and PFSP instances, and evaluates four agent-assisted workflows on
a small synthetic factory. A separate coding experiment tests implementation
of a new objective and hard constraint. The algorithm study covers a defined
model subset; the two agent studies examine configuration and native extension.
Model realism, search performance and agent-assisted use are complementary
parts of this programme.

\section{The APEX production-scheduling model}\label{sec:theory}

APEX extends the flexible job-shop model with production requirements drawn
from the authors' work across industrial settings and the research underlying
\textcite{Grumbach2023}. It addresses the detailed planning of discrete
production: alternative workplans, machines and personnel, shifts, preparation,
material availability and commitments from an existing plan. These are also
practical concerns when configuring manufacturing execution systems (MES) and
advanced planning and scheduling (APS) systems. Classical job shops and flow
shops are special cases; a permutation flow shop additionally requires a common
job order on every machine. Section~\ref{sec:data-methods} specifies how the
computational study isolates this classical core.

The model supports competing objectives as well as weighted and scaled
objectives arranged in lexicographic priority levels. Additional measures can
be implemented through the extension hooks described in
Section~\ref{sec:shared-evaluation}. Here the focus is on what a production
plan must represent and satisfy; the benchmark objectives are defined with
the experimental design.

An operation $i$ comprises main processing and any activated preparation,
follow-up or restart activities. An activity $a$ consists of ordered phases
$\phi$ using resources $r$. For an operation, $S_i$ is its original main start,
$E_i$ its main completion, and $R_i$ its product-ready time. For an activity,
$[S_a,E_a)$ is its execution envelope, including permitted pauses. Time is
measured in integer seconds within horizon $H$; intervals are half-open.
Unless stated otherwise, conditions apply only to present operations and
activated activities. The following relationships describe the implemented
semantics, rather than a standalone mixed-integer or constraint-programming
formulation.

\subsection{Orders, workplans and execution alternatives}

\paragraph{Production structure and quantities.}
Orders group jobs and their operations. Reusable workplans expand demand into
operations, dependencies and material requirements. A supplied maximum lot size
splits an order quantity into lots, including the final remainder. Lot quantities
are therefore inputs to scheduling after expansion; economic lot sizing is not
an additional optimisation decision. Releases, priorities, due dates and hard
deadlines describe the relevant operations and groups.

\paragraph{Alternative workplans and operation presence.}
A route choice can select between workplans containing different operations,
whereas an execution mode changes how an existing operation is performed.
Let $\mathcal P_g$ contain the alternatives of route choice $g$, and let binary
$z_{gp}$ select alternative $p$. An operation belonging to route choice $g(i)$
is present when the selected alternative contains it:
\begin{equation}
 \begin{gathered}
 \sum_{p\in\mathcal P_g}z_{gp}=1,\qquad
 A_i=\sum_{\substack{p\in\mathcal P_{g(i)}\\i\in p}}z_{g(i)p},\\
 A_i,z_{gp}\in\{0,1\}.
 \end{gathered}
 \label{eq:route-presence}
\end{equation}
Operations outside route choices have $A_i=1$. An operation may be shared by
alternatives of one choice, but cannot belong to different route choices.
The selected workplan activates its additional dependencies; dependencies
with inactive endpoints disappear. Selection must preserve operations already
started or protected by commitments.

\paragraph{Machines, personnel and execution modes.}
Each active operation selects exactly one permitted mode from $\mathcal M_i$:
\begin{equation}
 \begin{gathered}
 \sum_{m\in\mathcal M_i}x_{im}=A_i,\qquad x_{im}\in\{0,1\},\\
 x_{im}=0\quad\text{for an excluded mode}.
 \end{gathered}
 \label{eq:mode-selection}
\end{equation}
A mode specifies a primary resource, ordered phases and simultaneous resource
requirements. It may also change preparation, follow-up and material needs.
Thus alternatives can represent different machines, staffing arrangements or
production methods. A machine, qualified worker and fixture can be required
together; pooled resources can supply several capacity units. Eligibility is
expressed through the permitted modes, with resource identities fixed within
each selected phase. Every activated supporting activity similarly selects one
of its own modes. Fixing a resource can leave several modes available.

\paragraph{Work content and processing speed.}
For quantity $q_i$, phase work can combine a fixed amount and a per-unit amount:
\begin{equation}
 \begin{gathered}
 W_{im\phi}=W^{\mathrm{fixed}}_{im\phi}+q_iW^{\mathrm{unit}}_{im\phi},\\
 \sum_{s\in\mathcal S_{im\phi}}w_s=W_{im\phi}.
 \end{gathered}
 \label{eq:phase-work}
\end{equation}
Here $\mathcal S_{im\phi}$ contains the execution segments of the selected
mode's phase, and $w_s$ is the work completed in segment $s$. Work and elapsed
time differ: at constant positive rate $v$ without an interruption, phase
duration is $\lceil W_{im\phi}/v\rceil$. With varying rates, work accumulates
over calendar intervals using the designated rate resource; the default rate
is one. Integer time permits the final occupied second to be partly productive.
Material quantities supplied per unit in a workplan are scaled once during
expansion; the resulting operation-level quantities are totals.

\subsection{Resource capacity, calendars and activity timing}

\paragraph{Productive work and retained occupancy.}
An activity may occupy a machine even while processing is paused. Let
$u_{a\phi}(t)$ indicate a productive segment of phase $\phi$, and
$b_{a\phi r}$ its demand on resource $r$. Let $h_{ar}(t)$ denote all capacity
reserved for activity $a$, including required holds. Feasibility requires
\begin{equation}
 \begin{gathered}
 \begin{aligned}
 \sum_{a,\phi}b_{a\phi r}u_{a\phi}(t)
   &\le K_r^{\mathrm{processing}}(t),\\
 \sum_a h_{ar}(t)&\le K_r^{\mathrm{retention}}(t)
 \end{aligned}\\
 \text{for every resource $r$ and time $t$.}
 \end{gathered}
 \label{eq:capacity}
\end{equation}
The first calendar determines when work is allowed; the second determines
when capacity may remain occupied. All resources required for a phase must be
available simultaneously. If no retention calendar is supplied, nominal
physical capacity applies throughout the horizon.

Reservations are determined by the activity rules, not chosen independently
to make Equation~\eqref{eq:capacity} hold. The primary resource has constant
demand across the activity's complete envelope. A supporting resource is
reserved across a phase's envelope when retention is required, and only during
its productive segments otherwise. For example, a fixture can remain clamped
overnight while a worker is released; an unattended phase need not reserve
the worker used in an earlier phase.

\paragraph{Phase order, continuity and breaks.}
Consecutive phases within an activity satisfy
\begin{equation}
 \begin{gathered}
 S_{a,\phi+1}\ge E_{a\phi},\\
 S_{a,\phi+1}=E_{a\phi}\quad\text{if continuity is required}.
 \end{gathered}
 \label{eq:phase-order}
\end{equation}
Here $S_{a\phi}$ and $E_{a\phi}$ delimit phase $\phi$. A noninterruptible
phase has no internal gap. Calendar-resumable work can pause only while a
required processing calendar cannot support the work, including insufficient
capacity or zero rate. This permits shift breaks and overnight interruptions,
but not arbitrary preemption to process another operation. Any retained
resources must remain feasible throughout the pause.

\paragraph{Preparation, inspection and product readiness.}
Supporting activities have their own modes, resources and calendars. For an
unstarted operation, preparation and follow-up obey
\begin{equation}
 \begin{aligned}
 E_a&\le S_i\quad(a\in\mathcal A_i^{\mathrm{pre}}),\\
 S_a&\ge E_i\quad(a\in\mathcal A_i^{\mathrm{post}}).
 \end{aligned}
 \label{eq:activity-stages}
\end{equation}
Within either stage, a directed acyclic activity graph can allow parallel
branches; the default is serial list order. Let
$\mathcal A_i^{\mathrm{release}}\subseteq\mathcal A_i^{\mathrm{post}}$
contain the follow-up activities required before product release. Then
\begin{equation}
 R_i=\max\bigl(\{E_i\}\cup
        \{E_a:a\in\mathcal A_i^{\mathrm{release}}\}\bigr).
 \label{eq:product-readiness}
\end{equation}
Inspection on a separate resource may delay product release while the machine
already processes another operation. Conversely, cleanup may occupy the
machine after the product is ready. A single completion timestamp would lose
this distinction between product availability and resource occupation.

\paragraph{Dependencies, time lags and hard windows.}
A technological dependency from operation $i$ to $j$ uses product readiness:
\begin{equation}
 R_i+\ell_{ij}^{\min}\le S_j\le R_i+\ell_{ij}^{\max}.
 \label{eq:operation-lags}
\end{equation}
The upper bound is omitted when no maximum lag is declared. Minimum lags can
represent a required delay, while maximum lags limit how long a product may
wait. Within a supporting-activity stage the corresponding relation is
$E_a+\ell_{ab}^{\min}\le S_b\le E_a+\ell_{ab}^{\max}$.
Cross-operation links connect main-start and product-ready events; work needing
other cross-operation activity links can be represented as separate operations.

Release and completion bounds, together with the horizon, require
\begin{equation}
 S_i\ge r_i,\qquad R_i\le D_i,\qquad
 0\le S_a\le E_a\le H.
 \label{eq:time-bounds}
\end{equation}
For unstarted work, preparation also begins no earlier than its applicable
release boundary. Multiple windows are intersected to obtain effective bounds
$r_i$ and $D_i$. A hard deadline limits product readiness, whereas a soft due
date expresses a preference. Non-release-gating cleanup may finish after the
product deadline, but must remain within the horizon.

\paragraph{Sequence-dependent preparation and follow-up.}
The operation order $\pi_r$ on a primary resource determines which product
families meet. A family transition can activate preparation for the next
operation, follow-up for the previous one and separate charges. Longer family
patterns can trigger further activities. Initial resource states and terminal
cleanup are included; a declared transition table must cover every transition
used by a plan. These effects activate activities governed by the same work,
capacity and timing conditions as other activities. They can therefore express
more than a scalar changeover duration, such as cleaning on the machine
combined with inspection on a separate resource.

\subsection{Material availability, execution state and planning commitments}

\paragraph{Stock, receipts and intermediate products.}
Material is consumed at main start and produced at product readiness. For item
$k$, let $I_k^0$ be supplied opening stock, $c_{ik}$ consumption and $p_{ik}$
production under the selected mode. With $\mathcal U$ denoting active operations
not already started, the time-dependent balance is
\begin{equation}
 \begin{aligned}
 I_k(t)={}&I_k^0+\sum_{\tau\le t}\operatorname{receipt}_k(\tau)\\
         &+\sum_{i:R_i\le t}p_{ik}
          -\sum_{\substack{i\in\mathcal U\\S_i\le t}}c_{ik}\ge0.
 \end{aligned}
 \label{eq:material-balance}
\end{equation}
Inputs of started operations have already been consumed and are not deducted
again. The supplied stock and receipt ledger must consistently separate
existing inventory from the represented operation outputs. Mode-specific
material maps replace the corresponding operation-level maps.

An optional preparation step assigns declared supply sources $u$ to consumers:
\begin{equation}
 \begin{gathered}
 \sum_u q_{uik}=c_{ik},\qquad q_{uik}\ge0,\\
 \sum_i q_{uik}\le\operatorname{supply}_{uk}.
 \end{gathered}
 \label{eq:material-allocation}
\end{equation}
These conditions apply to the consumers included in allocation. Stock and
receipts retain their availability dates; allocation from a producing
operation adds an availability dependency. Allocation is deterministic and can
be recomputed after route or material-mode selection. The quantities
$q_{uik}$ are not independent global search variables, and allocation does
not create missing production or purchase orders.

\paragraph{Recorded execution and remaining work.}
Replanning preserves recorded execution, its resource reservations and the
selected mode. A running operation retains its original $S_i$; the new main
activity instead starts at $S_i^{\mathrm{resume}}$. With observation time
$t_i^{\mathrm{obs}}$ and supplied remaining work, continuation satisfies
\begin{equation}
 \begin{gathered}
 S_i^{\mathrm{resume}}\ge t_i^{\mathrm{obs}},\\
 \sum_{s\in\mathcal S_{i\phi}^{\mathrm{remaining}}}w_s
     =W_{i\phi}^{\mathrm{remaining}}.
 \end{gathered}
 \label{eq:remaining-work}
\end{equation}
Required restart activities precede resumption and use the same activity
constraints. Their stage boundary is the resumption start, not the historical
start in Equation~\eqref{eq:activity-stages}. The primary resource remains
reserved from observation until resumption, with these holds included in
$h_{ar}$ without double-counting restart occupancy. Future output becomes
available only after remaining work and release-gating follow-up finish.

\paragraph{Fixed decisions, resource order and freeze horizons.}
Commitments can separately fix a start, an execution mode or a primary resource:
\begin{equation}
 S_i=\bar S_i,\qquad x_{i\bar m}=1,\qquad
 \operatorname{primary}(i)=\bar r.
 \label{eq:fixed-decisions}
\end{equation}
Each equality applies only when that dimension is committed. Workplan and
supporting-activity choices can also be fixed. A relative resource-order
commitment requires $\operatorname{pos}_{\pi_r}(i)<\operatorname{pos}_{\pi_r}(j)$;
consecutive block members additionally require
$\operatorname{pos}_{\pi_r}(j)=\operatorname{pos}_{\pi_r}(i)+1$.
Consecutiveness excludes intervening operations, while allowing time gaps.
Resource order alone does not impose a product-release dependency: follow-up
on another resource may continue independently.

A freeze horizon selects operations by their baseline main start and protects
the specified decision dimensions. Cutoffs can differ by resource. Freezing
therefore need not fix every dimension or prohibit all new work before the
cutoff. At the group level, job and order completion are
$C_j=\max_{i\in\mathcal O_j}R_i$ and
$C_o=\max_{j\in\mathcal J_o}C_j$, where $\mathcal O_j$ contains active operations
of job $j$ and $\mathcal J_o$ contains the jobs of order $o$. Every required group
must have its represented work completed and satisfy its hard deadline.

\paragraph{Operating policies and extensions.}
Mandatory construction policies complement the physical constraints. They
restrict the next eligible operation--mode choice against the current
construction prefix before heuristic ranking. Campaign rules can require a
minimum amount of same-family work, idle-gap rules limit the next attainable
gap, and urgency rules prioritise imminent work under declared conditions.
Thresholds, exceptions and fallbacks are explicit. These are requirements on
the construction procedure, distinct from constraints on arbitrary intervals
of a completed schedule. Additional sequence effects, measures and hard checks
can be implemented and tested as extensions. Section~\ref{sec:search} explains
how construction and search use the common model; Section~\ref{sec:discussion}
addresses agent-assisted configuration and model extension.

\section{Construction and search}\label{sec:search}

\subsection{One evaluation path, different search objects}\label{sec:shared-evaluation}

\begin{figure*}[htbp]
\centering\includegraphics[width=\textwidth]{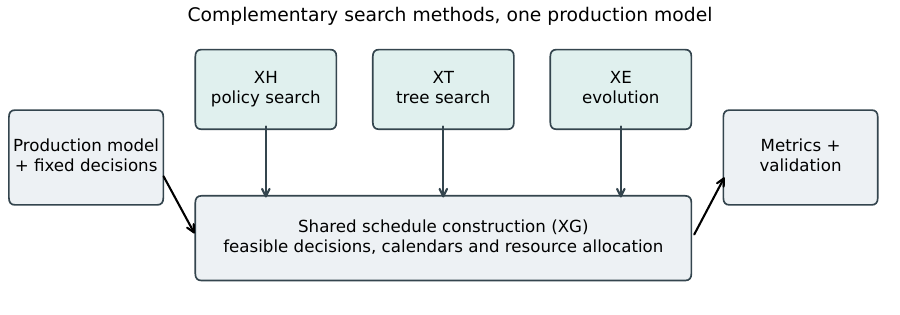}
\caption{Search components propose different decisions to a shared scheduling
core. The arrows indicate how proposed decisions are turned into checked
plans. Components can also be applied sequentially.}
\label{fig:architecture}
\end{figure*}

APEX uses four names in this paper: XG for greedy construction, XH for
hyper-heuristic policy search, XT for prefix-tree search and XE for direct
schedule evolution. The letters identify components, not separate physical
models. Figure~\ref{fig:architecture} shows the shared structure. Candidate
choices pass through dependency readiness, permitted task/mode decisions,
resource placement, objective evaluation and validation. Rejected candidates
cannot remain feasible alternatives through a favourable penalty score.

\paragraph{Customization hooks.}
The shared evaluation path exposes predefined hooks: points at which the
core calls compiled customization code. An application-specific module can
filter eligible operation/mode choices during construction, declare and
calculate additional objectives, and check completed schedules against
new rules. The interface also supports input preparation, additional
dispatch signals and sequence-dependent activities. Construction, objective
and validation hooks connect these extensions to the common evaluation
path used by XG, XH, XT and XE.

The hooks have distinct roles. Filters restrict the choices available to
construction; ranking signals order the remaining choices; exact objective
values compare completed plans. Final validation checks the added rules
alongside the core's resource, timing and commitment checks. This separates
application-specific decision logic from the search engines while allowing
it to change candidate admissibility and ranking. A coding agent can implement
and test the module; during planning, the scheduler executes its compiled
callbacks. Section~\ref{sec:native-extension-study} evaluates this path for
a new objective and a hard constraint.

XG constructs a complete schedule by repeatedly ranking permitted choices
using stage-specific dispatch signals, called Qs. Its initial route and mode choices
are heuristic. The same construction machinery can complete a schedule under
a supplied policy, decision prefix or operation-priority list. Complete
decoding, objective computation and checking are included in each candidate's
cost; prefix probes can add further work.

Representing a feasible production plan does not guarantee that construction
can reach it. The decoder appends work behind the previously constructed
occupancy on its primary resource and uses local placement decisions for
supporting activities. Material consumption follows a monotone time frontier.
These choices can narrow the reached set: a maximum-lag violation may reject a
candidate even when delaying its predecessor would permit a feasible plan.
Accepted output is checked against the model and commitments, with mandatory
policies verified against construction evidence. A failed construction or
exhausted search is therefore not a proof of infeasibility.

XH evolves construction policies, including weights and normalisation of
dispatch signals. It also searches unpinned routes and supporting-activity
modes; when material allocation is enabled, modes with different material
consumption or output become explicit search choices. Other unpinned main-mode
choices follow
the construction policy.
A policy is a reusable set of priorities for constructing a schedule, such
as how strongly to prefer short work or early completion. It is evaluated by
the complete schedule it produces. Here policy search adapts to the current
instance; it does not train a neural model. Pareto selection uses
nondominated ranks and crowding, concepts associated with NSGA-II
\parencite{Deb2002}. A separate incumbent is chosen by the configured
priority score.

XT explores a tree whose nodes fix decision prefixes. It selects branches
using an upper-confidence tree rule and completes prefixes with rollouts.
This follows the UCT selection principle of \textcite{Kocsis2006}.
The benchmark uses an equally weighted scalar score of the
normalised makespan and flowtime objectives for tree decisions and the single
incumbent. An external observer records all encountered nondominated
outcomes without steering the search.

\subsection{Direct schedule evolution}

XE evolves an operation-priority permutation together with mode, route and
supported conditional-activity choices. The permutation is a priority list,
not an unconstrained sequence of start times. A dependency-ready decoder
defers unavailable operations, enforces commitments and places permitted
activities. Different genomes can therefore decode to the same schedule.

Operators include insertion, swapping, job and resource block moves,
machine-mode and route changes, conditional-mode changes and a job-based
crossover. A discounted upper-confidence bandit allocates operator attempts.
Discounted UCB provides a precedent for adapting action estimates under
changing rewards \parencite{Garivier2008}; APEX's reward definition and
operator restrictions are specific implementation choices. In Pareto mode,
an offspring receives full credit for dominance over its reference parent,
partial credit for a distinct nondominated trade-off, and zero otherwise.
This reward does not directly maximise archive hypervolume.

XE filters operators against the actual parent and problem, skips unchanged
or previously proposed genomes before
evaluation, and uses binary-tournament parent selection. When both operator
kinds are applicable, it first chooses crossover with probability 0.5 and
otherwise mutation; the bandit selects an operator within that kind. Each
offspring receives one operator. This differs from applying crossover and
mutation in sequence.

Without a supplied seed, standalone XE constructs its own population. In the
benchmark configuration with population 16, its first candidate uses XG and
the remaining 15 use random construction. No XH or XT run precedes this
initialisation. A supplied schedule is instead converted to a genome, and
offspring build a population around it. A single supplied parent initially
permits mutation, while crossover requires distinct parents. A separate
incumbent preserves the best configured priority score even when population
truncation removes its genome.

\subsection{Combinations and budget accounting}

The benchmark evaluates XG, XH, XT and XE individually and combines them as
XHT, XHE, XTE and XHTE. Letters in a combination give the order of its
components: XHTE applies policy search, then tree search, then schedule
evolution. All phases share one total allowance. In a two-phase run,
the nominal allocation is 500 evaluations per phase; the three-phase run
uses 333, 333 and 334. Unused allowance carries forward. The tree receives
the best actually evaluated construction policy when preceded by XH.
XE receives one best-so-far validated incumbent, without turning inherited
machine choices into new hard commitments. The observational Pareto archive
does not seed later phases.

The operational improvement workflow can also supply several good, distinct
previous schedules to XE when that component is enabled. The experiment uses
one incumbent so that its handoff is clearly defined.

A complete-candidate allowance counts evaluated initial candidates and
offspring, including failed attempts. Skipped proposals in XE
consume time but no evaluation allowance. Internal prefix work is timed but
is not a separate complete-candidate evaluation. These distinctions matter
when comparing evaluation efficiency with wall-clock performance.

\section{Experimental design}\label{sec:data-methods}

\subsection{Classical problem classes and the tested core}

The study assesses the scheduling core on three established problem classes:
the job-shop problem (JSP), flexible job-shop problem (FJSP) and permutation
flow-shop problem (PFSP). Their public benchmark collections provide a common
basis for comparing sequencing and machine-assignment decisions. The
experiment deliberately isolates these decisions; it does not evaluate every
industrial requirement described in Section~\ref{sec:theory}.

For a job $j\in J$ with operation sequence $\mathcal O_j$, operation $i$ has
eligible machines $\mathcal E_i$ and processing time $p_{im}>0$. Choosing
machine $m(i)\in\mathcal E_i$ and start $S_i\ge0$ gives
\begin{equation}
 \begin{gathered}
 R_i=E_i=S_i+p_{i,m(i)},\\
 S_v\ge E_u\quad\text{for successive operations }(u,v).
 \end{gathered}
 \label{eq:classical}
\end{equation}
Operations assigned to the same machine cannot overlap. APEX represents this
FJSP core by one noninterruptible phase per operation, one mode per eligible
machine, unit processing rates and capacities, and full-horizon calendars.
All operations are mandatory; additional activities, material requirements
and industrial operating policies are absent. The mapping preserves the
classical constraints at the model's integer time resolution.

A single eligible machine per operation yields JSP. Requiring a common machine
route yields a flow shop; PFSP further requires the same job permutation on
every machine. The experiment explicitly enforces and independently checks
this common-permutation condition. Thus the benchmark tests three defined
restrictions of the general model, while preserving each class's established
problem semantics.

\subsection{Comparison set and public instances}

The numerical analysis covers 69 public instances: 19 JSP, 32 FJSP and
18 PFSP. This common comparison set was defined after execution by requiring
valid observations for all twelve methods and all three random seeds,
19, 42 and 73. The analysis is conditional on
this set, whose exact identifiers and dimensions are listed in
Table~\ref{tab:instances}. Repeated runs are nested within instances; they
are not additional independently sampled scheduling problems. All input processing
times remain unchanged. Releases are zero, machines have unit capacity and
processing is nonpreemptive. No dates or production extensions are added.

The JSP set combines 14 Taillard instances with \texttt{ft06}, attributed to
\textcite{Fisher1963}; \texttt{la01}, \texttt{la06} and \texttt{la16},
attributed to \textcite{Lawrence1984}; and \texttt{abz5}, attributed to
\textcite{Adams1988}. Acquisition uses the OR-Library job-shop collection
and its explicit family attributions \parencite{Beasley1990}. The FJSP set
contains 15 distributed Brandimarte instances, \texttt{mk01}--\texttt{mk15},
and 17 Behnke--Geiger work-centre instances
\parencite{Brandimarte1993,Behnke2012}, acquired from the SchedulingLab
distribution \parencite{SchedulingLab2026}. The extended Brandimarte
distribution is documented separately from the original publication in the
Behnke--Geiger report. The 18 PFSP instances and the additional JSP instances
come from Taillard's collections \parencite{Taillard1993}.

The JSP cases range from 6 to 50 jobs and 5 to 20 machines, with
36--1,000 operations. FJSP covers 10--50 jobs, 4--60 machines and
50--284 operations; its alternative machines introduce an additional
assignment decision. PFSP covers 20--100 jobs, 5--20 machines and
100--1,000 operations. These ranges describe the included set rather than
every size block in the source collections.

Raw source hashes, parsing provenance, generation seeds where supplied,
instance identifiers and resolved options are retained in the experiment
manifests. Historical bounds in source files remain metadata; the analysis
does not treat them as current best-known solutions or certified optima.

\subsection{External algorithms and discrete adapters}

The four external algorithms use a common multiobjective optimisation library
\parencite{Blank2020}. NSGA-II uses nondominated sorting and crowding
\parencite{Deb2002}; SPEA2 uses strength-based fitness, density estimation
and archive truncation \parencite{Zitzler2001}; MOEA/D works with neighbouring
scalar subproblems \parencite{Zhang2007}; and SMS-EMOA selects using
hypervolume contribution \parencite{Beume2007}. Population size is 100.
MOEA/D receives 100 evenly spaced two-objective reference directions.
Other library selection and survival settings are retained and exported.
In particular, SMS-EMOA uses the library's default batch offspring configuration;
we do not claim exact reproduction of the original steady-state procedure.

All four algorithms share the same discrete adapters within a problem class.
JSP uses a sequence of repeated job identifiers, decoded by taking the next
unscheduled operation of each selected job. FJSP adds machine choices indexed
by operation identity. These sequences use precedence-preserving order-based
crossover (POX), with uniform crossover on FJSP machine genes. POX and uniform machine
crossover have scheduling precedents \parencite{Phuang2018}. Mutation swaps
two sequence positions and, for FJSP, also attempts reassignment to another
eligible machine. PFSP uses a job permutation, two-point linear-order
crossover and swap mutation, as in the variation operators described by
\textcite{Chiang2010}. These are operator precedents; the complete published
search procedures and their parameters are not reproduced.

The crossover probability is 0.9 per mating, and the mutation probability is
1.0 per offspring. One mutation is a specified discrete move, not an
independent change of every gene. FJSP uniform machine crossover chooses
between parental genes with probability 0.5. Initialisation is random. No
local search or parameter tuning is added to the external algorithms.

External JSP/FJSP decoding inserts operations into the earliest feasible
machine gap. PFSP uses the usual completion recurrence for a fixed common
permutation. APEX retains its own placement decoder. Its benchmark-only PFSP
filter preserves the first-machine permutation on later machines and adds
validation cost. The comparison therefore includes decoder behaviour and
implementation overhead, rather than isolating the outer search method.

\subsection{Budgets, validation and runtime}

Each search configuration receives 1,000 complete-candidate evaluations;
XG performs one construction. Hybrid phases share this allowance
(Section~\ref{sec:search}). Parameters remain fixed throughout the comparison;
detailed settings are given in Appendix~\ref{sec:search-settings}.

Candidates undergo online feasibility checks. A separate checker audits every
exported nondominated schedule for feasibility and recomputes MS/FT. The
comparison uses all nondominated evaluated outcomes, including those outside
the final internal population.

Runtime is internal wall time, including search setup, construction, observation
and online validation. Data loading, final export and the separate audit are
excluded. Each algorithm uses one worker; eight independent runs execute
concurrently on the same host. Hardware and measurement details appear in
Appendix~\ref{sec:reproducibility}. Timing therefore characterises the measured
implementations under a common evaluation allowance.

\subsection{Objective normalisation and descriptive summaries}

For these inputs, job completion is $C_j=\max_{i\in\mathcal O_j}E_i$, and
the two minimised objectives are
\begin{equation}
 \MS=\max_{j\in J}C_j,\qquad \FT=\sum_{j\in J}C_j.
 \label{eq:objectives}
\end{equation}
Makespan measures when the workload finishes. Total job flowtime sums the time
each job spends in the system; with zero releases it equals the sum of job
completion times, not operation completion times. Product readiness and main
completion coincide here, and no additional activities extend makespan.
Tardiness is excluded because the selected instances do not supply due dates.

For an instance with $n$ jobs, define
$S=\sum_i\max_{m\in\mathcal E_i}p_{im}$. All implementations receive the
normalised objective vector
\begin{equation}
 f(x)=\left(\MS(x)/S,\;\FT(x)/(nS)\right).
 \label{eq:normalisation}
\end{equation}
The bounds depend only on the input, not on observed solver performance.
The tree and single-incumbent score weight both coordinates equally.
The objectives compete when a schedule that finishes all work earlier leaves
some individual jobs waiting longer. A Pareto archive retains alternatives
for which no evaluated schedule is at least as good in both objectives and
better in one. Hypervolume summarises both how good these alternatives are
and how much of the trade-off they cover. Formally, it is the area dominated by an observed
archive $A$ and bounded by the reference point $z=(1.1,1.1)$:
\begin{equation}
 \HV(A)=\lambda_2\!\left(\bigcup_{a\in A}
 [f_1(a),z_1]\times[f_2(a),z_2]\right),
 \label{eq:hv}
\end{equation}
where $\lambda_2$ denotes area \parencite{Beume2007}.

For each instance $i$, let $H_i^*$ be the largest hypervolume observed across
methods and seeds. The relative deficit of run $r$ is
$100(1-H_{ir}/H_i^*)$. The reference is the best observed run archive,
not the union of every archive and not a known Pareto front. Analogously,
MS and FT gaps use the lowest observed value on that instance. Best MS and
best FT within one run can belong to different schedules.

For each method and instance, we first take the median of the three seed
values, separately for each quality metric and runtime. We then average
these medians with equal weights for the 69 instances. Thus the overall
runtime is a mean of instance-level medians, not the median of all runs.
Class summaries apply the same procedure within JSP, FJSP or PFSP. An
additional appendix table gives equal weight to the three class means.
No significance tests are used.

Distribution plots show one seed median per instance, retaining every
observation. Boxes span the first and third quartiles, with the median
inside; whiskers reach the most extreme observed points within 1.5
interquartile ranges. Quantiles use linear interpolation. These plots show
heterogeneity across problems, not confidence intervals for an average.

Convergence curves reconstruct the nondominated archive from the saved
evaluation trace at each checkpoint. They retain the same instances at
every displayed checkpoint and use the final observed references above.
XG's single result is carried forward for comparison. Curves describe
progress within the recorded runs; they are not separate experiments
optimised for each smaller allowance. Post-run preference weights from
zero to one select solutions from the observed archive; they are not new
weight-specific optimisation runs.

\section{Computational results}\label{sec:results}

\subsection{Overall quality and computation}

All exported archive schedules in the 69-instance comparison pass independent
checks of feasibility and objective values.

\begin{table*}[htbp]
\centering\small\setlength{\tabcolsep}{5pt}
\caption{Overall results for the 69 instances: median over three seeds within
each instance, then an arithmetic mean with equal instance weights. Lower
is better. Quality is relative to the best observed result on the same
instance. Time is measured internal wall time. Bold marks column minima;
XG uses one construction and is excluded from the runtime minimum among
search methods, which each receive 1,000 evaluations.}
\label{tab:overall}
\begin{tabular}{lrrrr}
\toprule
Method & HV deficit (\%) & MS gap (\%) & FT gap (\%) & Time (s) \\
\midrule
XG & 3.498 & 18.405 & 14.587 & 0.035 \\
XH & 0.898 & 3.025 & 4.490 & 39.820 \\
XT & 1.364 & 6.910 & 6.976 & 35.606 \\
XE & 1.700 & 11.226 & 8.841 & 19.784 \\
XHT & 0.728 & 2.537 & 4.001 & 39.819 \\
XHE & 0.466 & 1.844 & \textbf{3.053} & 30.068 \\
XTE & 0.879 & 5.417 & 5.345 & 28.241 \\
XHTE & \textbf{0.457} & \textbf{1.717} & 3.329 & 33.129 \\
\midrule
NSGA-II & 2.886 & 23.490 & 26.473 & \textbf{0.542} \\
SPEA2 & 2.801 & 22.930 & 25.765 & 0.568 \\
MOEA/D & 2.248 & 18.998 & 22.665 & 0.715 \\
SMS-EMOA & 2.902 & 23.444 & 26.439 & 0.543 \\
\bottomrule
\end{tabular}
}
\end{table*}

XHTE and XHE provide the strongest overall archive quality
(Table~\ref{tab:overall}). Their mean relative hypervolume deficits are
0.457\% and 0.466\%, a difference of approximately 0.010 percentage points.
XHTE also has the smallest makespan gap, 1.717\%, whereas XHE has the
smallest flowtime gap, 3.053\%. The numerical lead in hypervolume therefore
does not identify a single configuration that is best on both objectives.
On paired instance medians, XHTE has the lower hypervolume deficit on
36 instances, XHE on 32, with one numerical tie.
MOEA/D has the lowest overall deficit among the four external algorithms,
at 2.248\%. Equal weighting of the three problem classes leaves XHTE and
XHE close, at 0.518\% and 0.520\%, respectively
(Table~\ref{tab:class-balanced}).

The one-construction XG reference takes 0.035\,s on the same aggregation
basis. Its mean makespan and flowtime gaps of 18.405\% and 14.587\%
show the quality difference between rapid construction and a longer search.
The low hypervolume deficit of 3.498\% should be interpreted alongside
these objective gaps because hypervolume measures an area in the normalised
objective space.

\subsection{Different problem classes favour different components}

\begin{figure*}[htbp]
\centering\includegraphics[width=\textwidth]{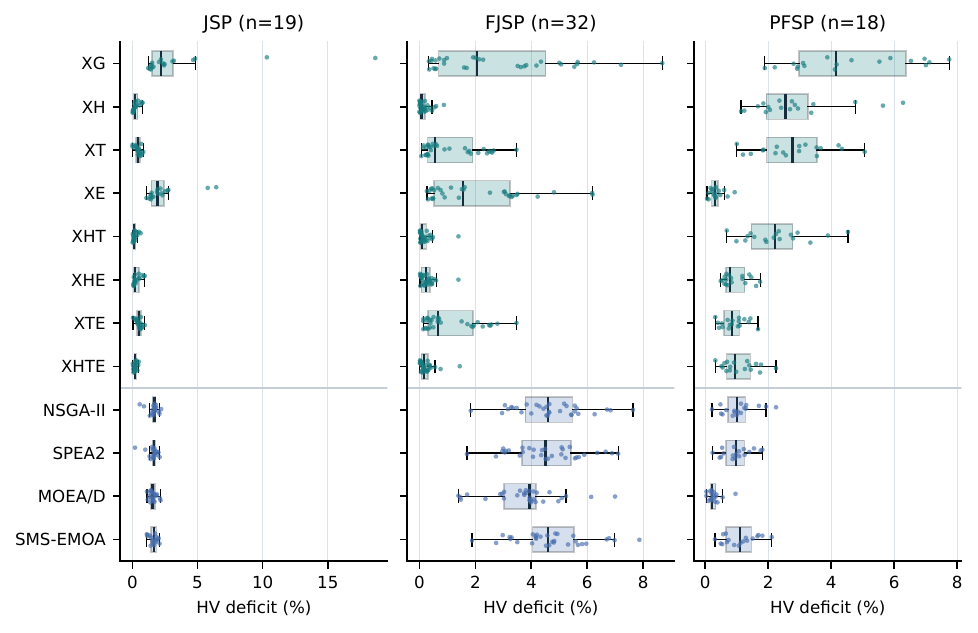}
\caption{Distribution of relative hypervolume deficits across the included
instances. Each point is one median over three seeds. Boxes show the
interquartile range and median; whiskers follow the 1.5-IQR rule. All
observations, including outliers, remain visible. Teal denotes APEX and
blue the external algorithms. Horizontal scales differ between panels.}
\label{fig:quality}
\end{figure*}

The aggregate comparison conceals substantial differences between problem
classes (Figure~\ref{fig:quality}). On JSP, XHT has the lowest mean
hypervolume deficit, 0.192\%, followed by XHTE at 0.224\% and XH at
0.271\%. On FJSP, XH leads at 0.171\%, followed by XHT at 0.193\%.
Their low deficits contrast with the broader distributions of standalone
XE and the external methods. These results make policy search a useful
component for the tested job-shop structures, while showing that additional
search phases do not improve every class-level average.

The ordering changes on PFSP. MOEA/D leads at 0.276\%, and standalone
XE is the strongest APEX configuration at 0.343\%. Their makespan gaps
are 0.957\% and 1.718\%, and their flowtime gaps are 1.529\% and
1.776\%, respectively. XE outperforms the APEX combinations on all
three class-mean quality measures here. A configuration's overall rank
therefore should not replace a problem-specific choice. Full class summaries,
objective distributions and per-instance values appear in the appendix.

\subsection{Quality and runtime are separate choices}

\begin{figure*}[htbp]
\centering\includegraphics[width=\textwidth]{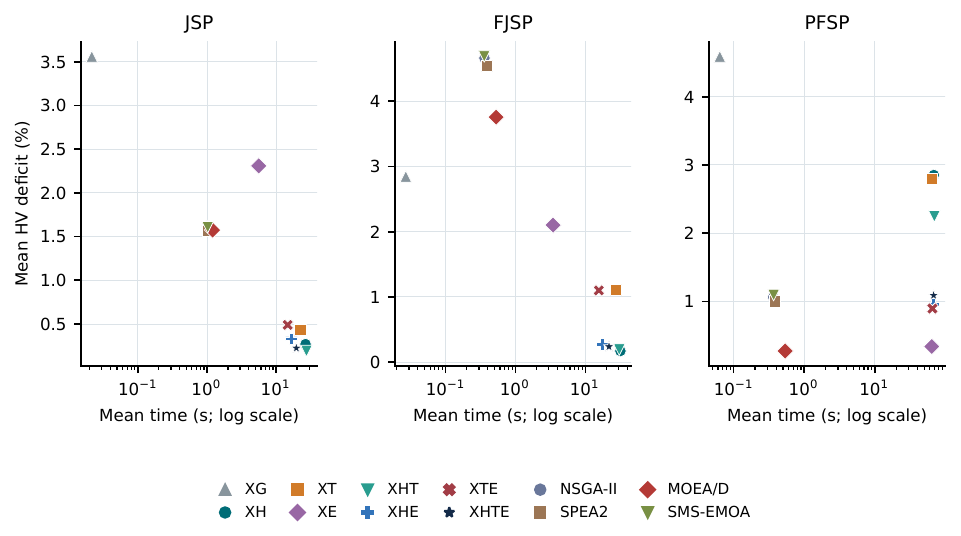}
\caption{Observed quality--time trade-off by problem class. Each point
combines the class means of instance-level seed medians. Lower and further
left indicate lower deficit and less measured time. Time uses a logarithmic
axis; XG is a single construction. The concurrent execution setting is
shared, but these are evaluation-budget comparisons, not equal-time runs.}
\label{fig:runtime}
\end{figure*}

The external methods require much less measured time at the common search
allowance (Figure~\ref{fig:runtime}). Their overall means range from
0.542\,s for NSGA-II to 0.715\,s for MOEA/D. The corresponding times
are 30.068\,s for XHE and 33.129\,s for XHTE. XHE thus combines
nearly the same aggregate archive quality as XHTE with a lower mean
flowtime gap and lower measured runtime. This makes it a relevant practical
alternative even though it does not occupy the first hypervolume rank.

PFSP illustrates why the joint comparison matters. XE has strong observed
quality, but its class-mean time is 63.810\,s, compared with 0.535\,s
for MOEA/D, which also has better class-mean quality. On JSP and FJSP,
APEX's better class-mean archive quality instead comes with greater
measured computation. The general construction and checking paths are
part of these implementation measurements; the experiment does not
attribute the difference to any one component.

Sequential combinations divide the fixed allowance between phases. Adding
XE can therefore coincide with a shorter total time because it replaces
some evaluations in earlier phases. For example, XHT uses 39.819\,s
on average, compared with 33.129\,s for XHTE. This compares allocations
of a shared allowance, rather than an unchanged search with additional work.

\subsection{Progress within the evaluation allowance}

\begin{figure*}[htbp]
\centering\includegraphics[width=\textwidth]{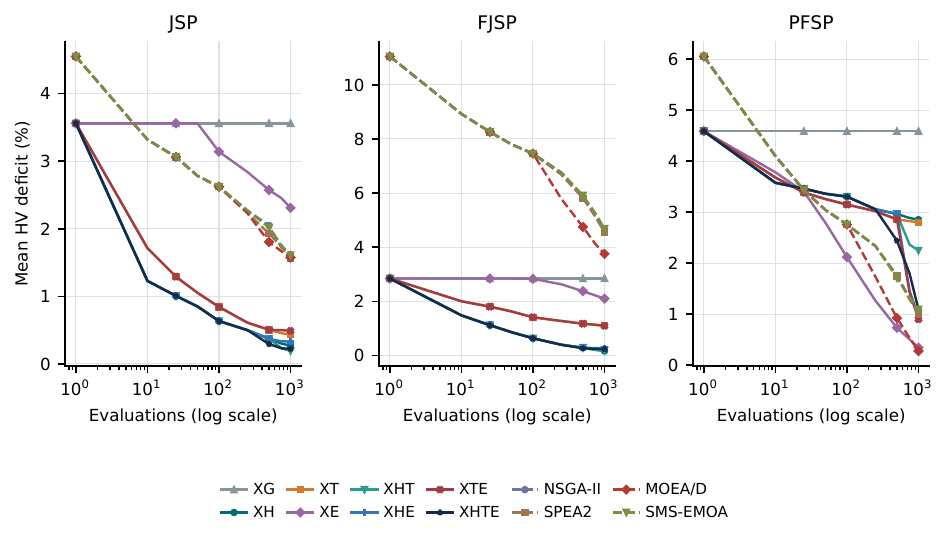}
\caption{Mean relative HV deficit of the observed archive at evaluation
checkpoints. Each class retains the same instances throughout, with seed
medians taken before averaging. Solid curves denote APEX; dashed curves
denote external methods. Checkpoints are connected for readability and
do not imply measurements between them. XG's result is carried forward
after its single construction. The reference is the final best observed
archive, not a known Pareto front.}
\label{fig:convergence}
\end{figure*}

Figure~\ref{fig:convergence} distinguishes final quality from how it is
reached. Policy-based configurations improve early on JSP and FJSP,
while standalone XE retains its initial archive for a larger part of the
allowance. On PFSP, XE and MOEA/D make substantial progress towards the
end of the displayed range. The combinations involving XE also improve
late, when their evolutionary phase receives the remaining allowance.
These traces are consistent with complementary behaviour of the components.

The observed class dependence supports selecting search configurations
according to the production structure and available computation.

\section{Agent-assisted operation and model extension}\label{sec:discussion}

The experiments in this section use OpenAI's GPT-6-astra
(model identifier \texttt{gpt-6-astra}, reasoning setting \texttt{xhigh}).
In the planning workflow, the agent serves as an interaction layer between
the human planner and APEX's algorithmic core: it translates planner requests
into model changes and tool calls, then explains the results. The core
constructs, evaluates and validates the schedules.

Agent assistance has two distinct roles in APEX: operating the planner
through existing model features and implementing additional semantics when
those features are insufficient. We evaluate them separately. The first
study covers rule changes, plan commitments, objective configuration and
what-if comparisons. The second requires the agent to write and test new
native objective and constraint code. Both start from synthetic data.

\begin{figure*}[htbp]
\centering
\includegraphics[width=\textwidth]{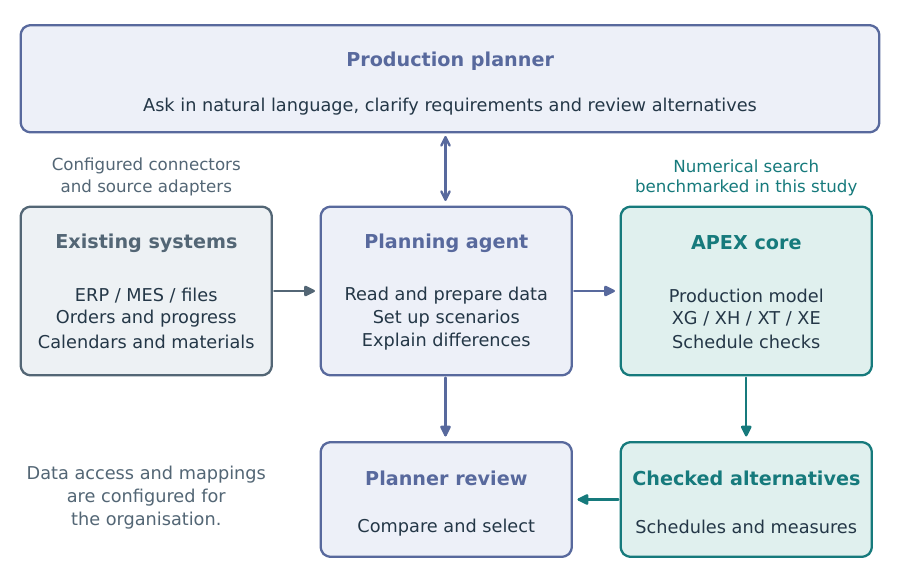}
\caption{From operational data to a reviewed planning alternative.
Adapters prepare the model, an agent coordinates changes, and APEX
constructs and checks schedules. The workflow experiment starts from a
prepared model; the coding experiment adds native semantics through
an isolated customization host.}
\label{fig:system-integration}
\end{figure*}

Configured connectors and adapters can supply orders, progress, calendars
and materials from enterprise resource planning (ERP), manufacturing execution
(MES), spreadsheets or databases. They reconcile identities, units and
production meaning before planning. Figure~\ref{fig:system-integration}
places the agent between these sources and the numerical core. This
interaction role resembles that studied by \textcite{Yuan2025}.

\paragraph{Operating an existing model.}
The workflow factory has six jobs with two successive non-preemptive
operations each. Stage one can use M1 or M2, and stage two M3 or M4;
processing times depend on the selected machine. All resources have unit
capacity and all releases are zero. Synthetic due dates, a hard deadline
and mode costs distinguish preferences from hard requirements. The
makespan-oriented baseline finishes at 4,800~s, with zero weighted
operation tardiness and total mode cost 30.

Two requests per category were repeated in three fresh contexts and
independent baseline copies. Full completion required the intended model
change, preservation of unrelated fields and the baseline, a physically
valid schedule for the current revision, and the requested final response
or comparison. A deterministic checker inspected saved models, schedules
and action traces. All 24 sessions passed, including all three repetitions
of every request (Table~\ref{tab:workflow-followup}). Median session time
was 44.5~s; each session used one XH search with 64 evaluations, whose
median internal time was 16.7~ms. Appendix~\ref{sec:workflow-protocol}
details the fixture, checks and measured interaction effort.

\begin{table*}[htbp]
\centering\small\setlength{\tabcolsep}{4pt}\renewcommand{\arraystretch}{1.13}
\caption{Configuration and planning workflows. Passes count complete
sessions. Request times and makespan (MS) are in seconds. Mode cost sums
synthetic production costs in arbitrary units (a.u.): four per operation
on M1 or M3 and one on M2 or M4, irrespective of duration. Each request
produced the same reported MS and cost in its three repetitions.
The baseline has MS 4,800~s and cost 30.}
\label{tab:workflow-followup}
\begin{tabularx}{\textwidth}{@{}lXrrr@{}}
\toprule
Workflow & Requested change & Passes & MS (s) & \shortstack[r]{Mode cost\\(a.u.)} \\
\midrule
Rules & M1 unavailable on $[0,2400)$ & 3/3 & 6,300 & 30 \\
 & J3-2 within $[7200,10800]$ & 3/3 & 9,300 & 30 \\
Plans & J2-1 on M2 at $3600$ & 3/3 & 8,400 & 30 \\
 & Freeze baseline starts before $1800$ & 3/3 & 4,800 & 30 \\
Objectives & Tardiness first, then makespan & 3/3 & 4,800 & 30 \\
 & Mode cost first, then makespan & 3/3 & 9,000 & 12 \\
What-if & M3 outage on $[0,3600)$; compare & 3/3 & 5,700 & 27 \\
 & Earlier due date and higher priority; compare & 3/3 & 4,800 & 30 \\
\bottomrule
\end{tabularx}
}
\end{table*}

\subsection{Adding planning rules}

The agent translated an M1 outage into a ban on processing and retention
from 0 to 2,400~s, and a separate request into a hard start/completion
window for J3-2. The resulting makespans were 6,300~s and 9,300~s.
The checks concerned the saved restrictions and their actual enforcement;
a longer plan can be the correct consequence of an added rule. These
requests use existing model features. A rule requiring new calculations
follows the implementation experiment in Section~\ref{sec:native-extension-study}.

\subsection{Changing plan commitments}

One request fixed J2-1 to M2 at exactly 3,600~s. The agent had to encode
both conditions as binding, yielding a makespan of 8,400~s. A second
request froze the machine and start of every operation beginning before
1,800~s in the baseline, then replanned the remaining work; makespan
remained 4,800~s. The checker verified the particular commitments,
not just the objective value. The original plan remained available
for comparison before accepting an alternative.

\subsection{Changing objectives}

The objective requests replaced the ranking hierarchy with either weighted
operation tardiness or total mode cost first, followed by makespan.
The tardiness-first result retained zero tardiness and the baseline
makespan. The cost-first result reduced cost from 30 to 12 while increasing
makespan to 9,000~s, exposing the requested trade-off. This tests correct
use of existing measures. Implementing a new formula additionally requires
an evaluator, integration into search fitness and independent verification,
as examined below.

\subsection{Comparing what-if scenarios}

The outage scenario made M3 unavailable from 0 to 3,600~s. The agent
correctly compared its validated plan with the baseline and reported a
900~s makespan increase. A second scenario advanced J6-2's soft due time
to 2,400~s and raised its priority to five while retaining the makespan
objective. The reported makespan change was zero, with weighted operation
tardiness of 12,000~s. This distinction matters: changing a due-date fact
does not turn it into a hard deadline or replace the chosen objective.

\subsection{Implementing new objectives and constraints}\label{sec:native-extension-study}

Configuration alone does not test whether the framework can accommodate
additional semantics. A second experiment therefore required the coding
agent to implement two new Rust customizations through the predefined
hooks in Section~\ref{sec:shared-evaluation}. The first adds priority-weighted squared operation
tardiness,
\begin{equation}
  f_{\mathrm{sq}}=\sum_{i:d_i\text{ defined}}w_i\,[\max(0,R_i-d_i)]^2,
\end{equation}
where $R_i$ is product-ready time, $d_i$ the soft due time and $w_i$ the
operation priority. Missing due times and zero priorities contribute zero.
The new measure is minimised first, with the existing makespan objective
retained second. Objective-declaration and metric-evaluation hooks connect
the new formula to search fitness, with its values recomputed during validation.

The second customization adds a hard balance rule. Each operation has a
nonnegative synthetic exposure value $e_i$, accumulated on its selected
primary workplace $r(i)$. For configured workplaces $\mathcal W$,
\begin{equation}
  \begin{gathered}
  L_r=\sum_{i:r(i)=r}e_i,\\
  \max_{r\in\mathcal W}L_r-\min_{r\in\mathcal W}L_r\leq\Delta.
  \end{gathered}
\end{equation}
Unused configured workplaces have load zero. The customization rejects
invalid input and uses candidate-filter and final-validation hooks.
The filter rejects a construction choice when conservative remaining-work
bounds show that no balanced completion is possible; temporary imbalance
alone is insufficient for rejection. Final validation enforces the inequality
on the complete schedule. These are new implementations for APEX, not claims that
squared penalties or workload balancing are new optimisation concepts.

Each coding session used a fresh GPT-6-astra context and received the
extension instructions, interface declarations and one public example,
with at most four compile-and-example calls. Two
additional data variants and the acceptance checker were concealed. The
checker independently recomputed schedule feasibility and the new formula
or inequality, examined preservation of existing model fields, and tested
invalid inputs and corrupted outputs. Appendix~\ref{sec:native-extension-protocol}
documents the fixtures, time allowances and controls.

\begin{table*}[htbp]
\centering\small\setlength{\tabcolsep}{4pt}\renewcommand{\arraystretch}{1.13}
\caption{Acceptance of submitted native implementations, including the
additional constraint attempt. Counts give accepted final modules over
submitted final modules. Checks are acceptance assertions per module.
Median time covers sessions that produced a final module and includes code
generation, compilation and public-example feedback. The seven attempts,
including one without a submission, are documented in
Appendix~\ref{sec:native-extension-protocol}.}
\label{tab:native-extension}
\begin{tabularx}{\textwidth}{@{}Xrrr@{}}
\toprule
New implementation & Accepted/submitted & Checks & Median time (s) \\
\midrule
Squared-tardiness objective & 3/3 & 68 & 198.0 \\
Exposure-balance constraint & 3/3 & 61 & 249.6 \\
\bottomrule
\end{tabularx}
}
\end{table*}

All six submitted final implementations passed their complete acceptance
suites: three for the objective and three for the constraint
(Table~\ref{tab:native-extension}). Seven attempts produced these six
submissions; one constraint session reached its time limit without submitting
code. On the public objective example, every
repeat reduced the independently calculated squared penalty from 13 to 7,
while makespan increased from 6 to 7~s. For the hard rule, the workplace
exposure difference fell from four points to zero, with makespan increasing
from 3 to 4~s. These are the requested trade-offs: a new preference changes
ranking, whereas a new hard rule excludes previously acceptable plans.

The experiment executed the generated code through an isolated library
host. All six final modules are preserved with the study records;
the scheduling core and its manifests were unchanged. The objective uses
the existing due-date, slack and priority Qs; no new Q was synthesized.
The results demonstrate the extension path on both requests, combining
compiler feedback with external acceptance checks.

\subsection{Towards a digital twin for production decisions}

The four workflows operate on a planning snapshot. A digital twin additionally
requires a continuing relationship with its target system. ISO/IEC~30173
describes a digital representation whose data connections keep physical and
digital states aligned at a rate appropriate to their purpose
\parencite[clause~3.1.1]{ISO2023}. For manufacturing, the ISO~23247 framework
centres on a representation suited to its task and synchronised with
observable manufacturing elements \parencite[Section~2.1]{Shao2021}.
A scheduling twin can therefore focus on orders, resources, materials and
execution status without reproducing every physical detail of a machine.

\begin{figure*}[!t]
\centering
\includegraphics[width=\textwidth]{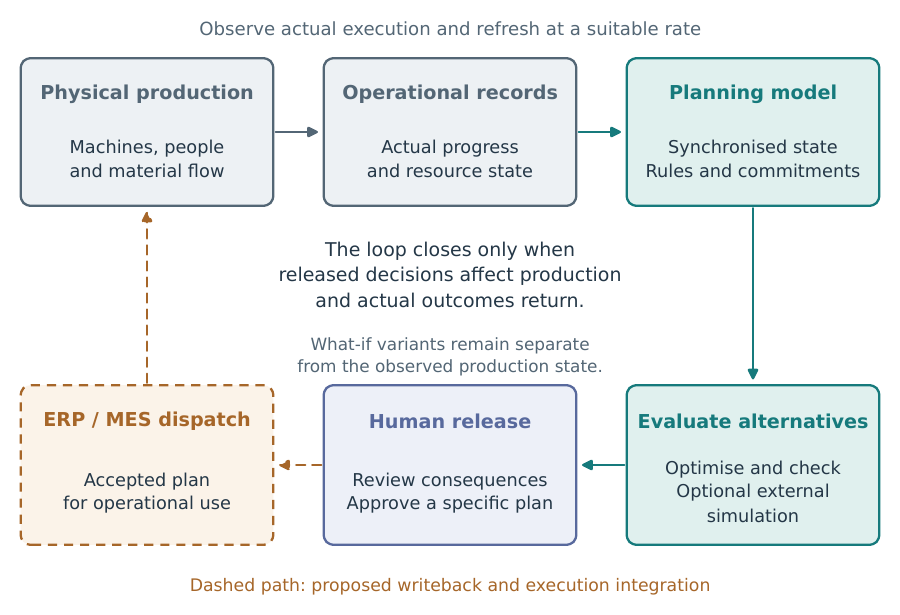}
\caption{Proposed integration of APEX into a digital twin for production
decisions. Observed state supports planning and scenario comparison; human
release connects an accepted alternative to execution. Dashed arrows mark
the additional writeback and dispatch path, which is not implemented in
the current planning workflow.}
\label{fig:digital-twin-loop}
\end{figure*}

\paragraph{Releasing plans into production.}
Figure~\ref{fig:digital-twin-loop} shows a proposed closed-loop integration.
Stable identities connect source records to the planning model; progress,
machine availability and material changes refresh the observed state.
Hypothetical alternatives remain separate. An authorised planner reviews
a candidate against an approved model and releases a specific plan version.
A deployment-specific return path would
check that the source state is still applicable and transfer the accepted
changes to the ERP or MES governing execution. Subsequent execution records
would update the model and initiate the next planning cycle.

Human release is compatible with the NIST scheduling and routing use case,
which allows schedules to reach production automatically or through user
recommendations \parencite[Section~4.2]{Shao2021}. The proposed loop still
requires execution feedback; writing a schedule into an ERP record alone
does not establish it.

\paragraph{Approving model and agent changes.}
Agent-assisted refinement introduces a second approval object: the rules
under which future plans are generated. Changes to constraints, objectives
or decision logic can alter which schedules are feasible or preferred,
even when they require only configuration. A company therefore needs to
define which adjustments fall within an approved scope and which require
renewed model approval. It also needs to decide which actions an agent may
perform independently and who can authorise changes to that authority.

Designing this process remains an open organisational and technical question.
A proposed approval would combine tests of the implementation with a
production expert's review of the intended meaning and effects on
representative scenarios. Passing technical checks alone does not establish
that a new rule expresses the company's requirements. Responsibilities,
required evidence, version records and a way to restore an earlier approved
model would need to fit the local planning process. Model approval would
permit use of the revised rules; release of a resulting production plan
would remain a separate decision.

\section{Limitations and next evidence steps}\label{sec:limitations}

\paragraph{Comparison set and descriptive evidence.}
The reported findings are conditional on the listed 69 public instances.
The set spans several sizes and problem families but is not a representative
random sample of industrial production problems. Requiring completion by every
method and seed conditions the comparison on instances amenable to all tested
implementations; it does not measure completion reliability. Its three classes have
different numbers of instances, so the overall mean gives more weight to
FJSP. Equal-class aggregation is a supplementary perspective on the same
observations, not independent confirmation. Three seeds give a limited
description of stochastic variation. The reported rankings do not establish
statistically reliable differences or consistent superiority on unseen
instances; the small aggregate difference between XHTE and XHE is a
particularly clear example.

\paragraph{Comparison scope and time.}
The baselines are documented library algorithms with discrete adapters,
not a survey of fully tuned scheduling state of the art. Specialised CP,
CP-based neighbourhood search and learned scheduling policies discussed in
Section~\ref{sec:related-work} are not included as baselines. Population sizes
and decoders differ. Equal complete-candidate counts do not equalise prefix
probes, rejected proposals, validation work or runtime. Eight concurrent
single-worker processes share the host, whose heterogeneous processors
and competing workloads affect wall time. The measurements describe this
execution setting rather than isolated algorithmic cost. Matched-time
experiments on an otherwise idle host are needed to compare quality for
a fixed computational deadline. The 1,000-evaluation allowance is also
modest for the larger included instances. Sequential combinations change both
the starting schedule and the allocation across phases, so they do not isolate
each component's causal contribution. Multiple-seed handoffs to XE were not
evaluated.

\paragraph{Metrics and references.}
Observed best values are not optima. Hypervolume depends on scaling and the
reference point; the serial-work normalisation may compress differences
when attainable completion times are much smaller than the serial bound.
Sub-percent hypervolume deficits must therefore not be read as sub-percent
distance to an optimal schedule. Separate MS/FT gaps and per-instance
archives provide complementary information. The two objective minima may
belong to different schedules. Post-run weights describe choices within
an observed archive, not performance after rerunning a method for a new
preference. The classical benchmark has no due-date data and therefore
supports no inference about tardiness; the separate workflow fixture supplies
synthetic due dates.

\paragraph{Model and deployment reach.}
The model mapping establishes representation of the classical subset at the
implemented time resolution, not search completeness. The core can miss
feasible plans, reconstructs candidates in full, and does not implement
arbitrary preemption, unrestricted cross-task internal activity graphs or
stochastic simulation. New native semantics require code and tests; a new
metric does not automatically supply an effective heuristic. The cited
earlier production studies do not validate this implementation. Extended-model
performance, human-centred objectives and industrial outcomes need separately
scoped evidence. The proposed approval processes for plans and model changes
have not been evaluated in industrial operation. Policy reuse under changed
conditions, Monte Carlo throughput and resulting robustness gains remain to
be measured. The snapshot studies establish neither a deployed feedback loop
nor conformity with digital-twin standards.

\paragraph{Synthetic agent-study scope.}
The configuration study covers one small factory and eight precisely specified
requests, each repeated three times. Repetition measures consistency on
these tasks, not reliability across production settings. The requests were
known during guidance preparation; this is a test on familiar requests,
not a held-out evaluation.
Clarification and conflict handling were excluded from its four-category
protocol. The checker verifies the declared task semantics, but does not
establish that an underspecified real-world request has been interpreted
correctly. The separate coding study adds only two precisely specified native
behaviours, tested on small, designed fixtures. Compiler feedback and public
examples are part of its support; concealed checks do not make it an
independent confirmation of general extension reliability. The additional
constraint attempt was selected after a timeout and uses a fresh context with
a longer time allowance. The reported acceptance counts concern submitted
final modules, not completion across all attempts under a common time limit.
The study uses an isolated
library host, not deployment into the product service. New dispatch functions,
mathematical middleware and live ERP/MES integration were not tested.
The agent-study evidence remains a local package, separate from the publicly
available algorithm benchmark. Session time depends on the model service and execution environment; there
is no human-time or industrial-productivity comparison. Next tests should
use unseen requests and instances, ambiguous instructions and interacting
changes, followed by planner studies and operational feedback measurements.

\section{Conclusion}\label{sec:conclusion}

APEX combines complementary search methods with a production model shaped by
industrial planning requirements. On the 69-instance comparison set,
XHTE and XHE achieve closely matched aggregate trade-off quality. XHTE
has the smallest mean hypervolume deficit and makespan gap, while XHE
has the smallest flowtime gap and a lower measured runtime. The
class-specific hypervolume leaders are XHT for JSP, XH for FJSP and
MOEA/D for PFSP; standalone XE is the strongest APEX configuration on
PFSP. These differences support choosing components for the problem
structure rather than assuming that the longest combination is always best.
The external implementations require substantially less measured time at
the common candidate allowance, making runtime improvement and matched-time
comparisons priorities for further work.

The separate workflow experiment shows that an agent can operate the model
correctly on the eight tested requests: all 24 sessions completed the rule,
plan, objective or what-if task while preserving the baseline. The evidence
concerns supported configuration and planning operations on a small synthetic
factory. A separate coding experiment extends that
evidence beyond configuration: all six submitted final implementations of the new
squared-tardiness objective or hard exposure-balance rule passed the prescribed
independent checks. The generated customizations changed schedule
ranking or admissibility while leaving the core unchanged.

The broader contribution is a generic, extensible production model coupled
to fast schedule construction and agent-assisted use and refinement.
Workplans, resources, materials and commitments share one representation
across the search methods. Agents support configuration and scenario work;
coding agents help implement and test new constraints and objectives.
On the classical benchmark instances, a single XG construction took 0.035~s on
average across instance-level seed medians (Table~\ref{tab:overall}). XH
policy search can invest additional computation in the dispatch priorities
that guide this construction. Once selected, a policy can be supplied to
the same greedy constructor, separating the search for useful priorities
from their subsequent application. This combination provides a basis for
rapid planning within the model's supported production constraints.

A natural next application is Monte Carlo simulation of production under
uncertainty, building on simulation-based assessment of robust and stable
schedules \parencite{Grumbach2024b}. Sampled processing times, material
arrivals, machine failures and repair durations would generate alternative
execution trajectories. XG could construct or repair schedules at simulated
decision points, using priorities prepared by XH. Repeated application
could spread the cost of policy search across many such decisions while
retaining a common model of what production permits.

Testing this application requires evaluating candidate plans or policies
across many trajectories for expected performance, adverse outcomes and
disruption of existing commitments. Reactive decisions must use only the
information available at the simulated decision time, and independent
scenarios should test the selected candidates.

Alongside simulation, an industrial field study should test APEX with
production planners, live operational data and execution feedback. A central
open question is how companies organise two distinct approvals: releasing a
selected plan into production and accepting changes to the model, objectives,
decision logic or the agent's permitted actions. Technical validation supports
these decisions but cannot replace approval of their operational meaning.
A progression from shadow operation to supervised use would allow the study
to examine model fit, approval effort, planner workload and realised production
outcomes, while developing an approval process suited to the company.

\FloatBarrier
\printbibliography

@article{Grumbach2023,
  author = {Felix Grumbach and Nour Eldin Alaa Badr and Pascal Reusch and Sebastian Trojahn},
  title = {A Memetic Algorithm With Reinforcement Learning for Sociotechnical Production Scheduling},
  year = {2023},
  journal = {IEEE Access},
  volume = {11},
  pages = {68760--68775},
  doi = {10.1109/ACCESS.2023.3292548},
}

@article{Vollenkemper2023,
  author = {Lukas Vollenkemper and Felix Grumbach and Martin Kohlhase and Pascal Reusch},
  title = {Humanzentrierte Ablaufplanung von Montagelinien},
  year = {2023},
  subtitle = {Plug and Play -- effiziente Algorithmen minimieren Belastung an Transferstraßen},
  journal = {wt Werkstattstechnik online},
  volume = {113},
  number = {4},
  pages = {158--163},
  doi = {10.37544/1436-4980-2023-04-58},
}

@thesis{Grumbach2024,
  author = {Felix Grumbach},
  title = {Feldsynchrone Ablaufplanung dynamischer Fertigungsprozesse mit Techniken des maschinellen Lernens},
  year = {2024},
  institution = {Hochschule Anhalt},
  type = {Doctoral dissertation},
  doi = {10.25673/115290},
}

@article{Deb2002,
  author = {Kalyanmoy Deb and Amrit Pratap and Sameer Agarwal and T. Meyarivan},
  title = {A Fast and Elitist Multiobjective Genetic Algorithm: {NSGA-II}},
  year = {2002},
  journal = {IEEE Transactions on Evolutionary Computation},
  volume = {6},
  number = {2},
  pages = {182--197},
  doi = {10.1109/4235.996017},
}

@report{Zitzler2001,
  author = {Eckart Zitzler and Marco Laumanns and Lothar Thiele},
  title = {{SPEA2}: Improving the Strength Pareto Evolutionary Algorithm},
  year = {2001},
  institution = {ETH Zurich, Computer Engineering and Networks Laboratory},
  number = {103},
  doi = {10.3929/ethz-a-004284029},
}

@article{Zhang2007,
  author = {Qingfu Zhang and Hui Li},
  title = {{MOEA/D}: A Multiobjective Evolutionary Algorithm Based on Decomposition},
  year = {2007},
  journal = {IEEE Transactions on Evolutionary Computation},
  volume = {11},
  number = {6},
  pages = {712--731},
  doi = {10.1109/TEVC.2007.892759},
}

@article{Beume2007,
  author = {Nicola Beume and Boris Naujoks and Michael Emmerich},
  title = {{SMS-EMOA}: Multiobjective selection based on dominated hypervolume},
  year = {2007},
  journal = {European Journal of Operational Research},
  volume = {181},
  number = {3},
  pages = {1653--1669},
  doi = {10.1016/j.ejor.2006.08.008},
}

@article{Blank2020,
  author = {Julian Blank and Kalyanmoy Deb},
  title = {pymoo: Multi-Objective Optimization in Python},
  year = {2020},
  journal = {IEEE Access},
  volume = {8},
  pages = {89497--89509},
  doi = {10.1109/ACCESS.2020.2990567},
}

@inproceedings{Kocsis2006,
  author = {Levente Kocsis and Csaba Szepesvári},
  title = {Bandit Based Monte-Carlo Planning},
  year = {2006},
  booktitle = {Machine Learning: ECML 2006},
  pages = {282--293},
  doi = {10.1007/11871842_29},
}

@online{Garivier2008,
  author = {Aurélien Garivier and Eric Moulines},
  title = {On Upper-Confidence Bound Policies for Non-Stationary Bandit Problems},
  year = {2008},
  eprint = {0805.3415},
  eprinttype = {arxiv},
  url = {https://arxiv.org/abs/0805.3415},
  urldate = {2026-09-24},
}

@article{Phuang2018,
  author = {Ajchara Phu-ang},
  title = {Discrete Differential Evolution Algorithm with the Fuzzy Machine Selection for Solving the Flexible Job Shop Scheduling Problem},
  year = {2018},
  journal = {International Journal of Networked and Distributed Computing},
  volume = {7},
  number = {1},
  pages = {11--19},
  doi = {10.2991/ijndc.2018.7.1.2},
}

@inproceedings{Chiang2010,
  author = {Tsung-Che Chiang and Li-Chen Fu},
  title = {An improved multiobjective memetic algorithm for permutation flow shop scheduling},
  year = {2010},
  booktitle = {2010 IEEE Congress on Evolutionary Computation},
  pages = {1--8},
  doi = {10.1109/CEC.2010.5586141},
}

@incollection{Fisher1963,
  author = {H. Fisher and G. L. Thompson},
  title = {Probabilistic learning combinations of local job-shop scheduling rules},
  year = {1963},
  booktitle = {Industrial Scheduling},
  publisher = {Prentice-Hall},
  editor = {J. F. Muth and G. L. Thompson},
  pages = {225--251},
  url = {https://people.brunel.ac.uk/~mastjjb/jeb/orlib/files/jobshop1.txt},
}

@report{Lawrence1984,
  author = {S. Lawrence},
  title = {Resource constrained project scheduling: an experimental investigation of heuristic scheduling techniques (Supplement)},
  year = {1984},
  institution = {Graduate School of Industrial Administration, Carnegie-Mellon University},
  url = {https://people.brunel.ac.uk/~mastjjb/jeb/orlib/files/jobshop1.txt},
}

@article{Adams1988,
  author = {Joseph Adams and Egon Balas and Daniel Zawack},
  title = {The Shifting Bottleneck Procedure for Job Shop Scheduling},
  year = {1988},
  journal = {Management Science},
  volume = {34},
  number = {3},
  pages = {391--401},
  doi = {10.1287/mnsc.34.3.391},
}

@article{Brandimarte1993,
  author = {Paolo Brandimarte},
  title = {Routing and scheduling in a flexible job shop by tabu search},
  year = {1993},
  journal = {Annals of Operations Research},
  volume = {41},
  pages = {157--183},
  doi = {10.1007/BF02023073},
}

@article{Taillard1993,
  author = {Éric Taillard},
  title = {Benchmarks for basic scheduling problems},
  year = {1993},
  journal = {European Journal of Operational Research},
  volume = {64},
  number = {2},
  pages = {278--285},
  doi = {10.1016/0377-2217(93)90182-M},
}

@article{Beasley1990,
  author = {J. E. Beasley},
  title = {{OR-Library}: Distributing Test Problems by Electronic Mail},
  year = {1990},
  journal = {Journal of the Operational Research Society},
  volume = {41},
  number = {11},
  pages = {1069--1072},
  doi = {10.1057/jors.1990.166},
}

@online{SchedulingLab2026,
  author = {{SchedulingLab}},
  title = {Flexible Job-shop Instances},
  note = {Dataset repository},
  url = {https://github.com/SchedulingLab/fjsp-instances},
  urldate = {2026-09-24},
}

@report{Behnke2012,
  author = {Dennis Behnke and Martin Josef Geiger},
  title = {Test Instances for the Flexible Job Shop Scheduling Problem with Work Centers},
  year = {2012},
  institution = {Helmut-Schmidt University},
  number = {RR-12-01-01},
  url = {https://d-nb.info/1023241773/34},
}

@incollection{Bakopoulos2024,
  author = {Emmanouil Bakopoulos and Vasilis Siatras and Panagiotis Mavrothalassitis and Nikolaos Nikolakis and Kosmas Alexopoulos},
  title = {Digital-Twin-Enabled Framework for Training and Deploying {AI} Agents for Production Scheduling},
  year = {2024},
  booktitle = {Artificial Intelligence in Manufacturing},
  publisher = {Springer},
  editor = {John Soldatos},
  pages = {147--179},
  doi = {10.1007/978-3-031-46452-2_9},
}

@inproceedings{Xia2024,
  author = {Yuchen Xia and Daniel Dittler and Nasser Jazdi and Haonan Chen and Michael Weyrich},
  title = {{LLM} experiments with simulation: Large Language Model Multi-Agent System for Simulation Model Parametrization in Digital Twins},
  year = {2024},
  booktitle = {2024 IEEE 29th International Conference on Emerging Technologies and Factory Automation (ETFA)},
  pages = {1--4},
  doi = {10.1109/ETFA61755.2024.10710900},
}

@article{Yuan2025,
  author = {Zhaolin Yuan and Ming Li and Chang Liu and Fangyuan Han and Haolun Huang and Hong-Ning Dai},
  title = {Chat with {MES}: {LLM}-driven user interface for manipulating garment manufacturing system through natural language},
  year = {2025},
  journal = {Journal of Manufacturing Systems},
  volume = {80},
  pages = {1093--1107},
  doi = {10.1016/j.jmsy.2025.02.008},
}

@article{Wang2025,
  author = {Zelong Wang and Chenhui Wan and Jie Liu and Xi Zhang and Haifeng Wang and Youmin Hu and Zhongxu Hu},
  title = {{MASC}: Large language model-based multi-agent scheduling chain for flexible job shop scheduling problem},
  year = {2025},
  journal = {Advanced Engineering Informatics},
  volume = {67},
  pages = {103527},
  doi = {10.1016/j.aei.2025.103527},
}

@article{May2026,
  author = {Marvin Carl May and Shady Salama and Johannes Pflüger and Toshiya Kaihara},
  title = {{LLM} actor-critic based dispatching rule generation for dynamic job shop scheduling},
  year = {2026},
  journal = {CIRP Annals},
  volume = {75},
  number = {1},
  pages = {601--605},
  doi = {10.1016/j.cirp.2026.04.007},
}

@article{Korth2026,
  author = {Merlin Korth and Martin Benfer and Gisela Lanza},
  title = {Multi-agentic production planning utilising simulation and optimisation},
  year = {2026},
  journal = {CIRP Annals},
  volume = {75},
  number = {1},
  pages = {595--599},
  doi = {10.1016/j.cirp.2026.04.027},
}

@report{Ye2026,
  author = {Wei Ye and Marvin Carl May and József Váncza and Xingyu Li},
  title = {Human-in-the-Loop Multi-Objective Manufacturing Scheduling Optimization with {LLM} Agents},
  year = {2026},
  institution = {SSRN},
  type = {Preprint},
  number = {7039079},
  doi = {10.2139/ssrn.7039079},
}

@report{ISO2023,
  author = {{{ISO/IEC}}},
  title = {{ISO/IEC} 30173:2023: Digital twin---Concepts and terminology},
  year = {2023},
  institution = {International Organization for Standardization and International Electrotechnical Commission},
  type = {International Standard},
  url = {https://www.iso.org/standard/81442.html},
}

@report{Shao2021,
  author = {Guodong Shao},
  title = {Use Case Scenarios for Digital Twin Implementation Based on {ISO} 23247},
  year = {2021},
  institution = {National Institute of Standards and Technology},
  type = {NIST Advanced Manufacturing Series},
  number = {400-2},
  doi = {10.6028/NIST.AMS.400-2},
}

@article{Grumbach2024b,
  author = {Felix Grumbach and Anna Müller and Pascal Reusch and Sebastian Trojahn},
  title = {Robust-stable scheduling in dynamic flow shops based on deep reinforcement learning},
  year = {2024},
  journal = {Journal of Intelligent Manufacturing},
  volume = {35},
  number = {2},
  pages = {667--686},
  doi = {10.1007/s10845-022-02069-x},
}

@inproceedings{Ahmaditeshnizi2024,
  author = {Ali Ahmaditeshnizi and Wenzhi Gao and Madeleine Udell},
  title = {{OptiMUS}: Scalable Optimization Modeling with ({MI}){LP} Solvers and Large Language Models},
  year = {2024},
  booktitle = {Proceedings of the 41st International Conference on Machine Learning},
  publisher = {PMLR},
  volume = {235},
  pages = {577--596},
  url = {https://proceedings.mlr.press/v235/ahmaditeshnizi24a.html},
}

@article{KjellsdotterIvert2011,
  author = {Kjellsdotter Ivert, Linea and Jonsson, Patrik},
  title = {Problems in the onward and upward phase of {APS} system implementation: Why do they occur?},
  year = {2011},
  journal = {International Journal of Physical Distribution \& Logistics Management},
  volume = {41},
  number = {4},
  pages = {343--363},
  doi = {10.1108/09600031111131922},
}

@article{Eriksson2022,
  author = {Kristina M. Eriksson and Linnéa Carlsson and Anna Karin Olsson},
  title = {To digitalize or not? Navigating and merging human- and technology perspectives in production planning and control},
  year = {2022},
  journal = {The International Journal of Advanced Manufacturing Technology},
  volume = {122},
  number = {11--12},
  pages = {4365--4373},
  doi = {10.1007/s00170-022-09874-x},
}

@misc{Lan2025,
  author = {Leon Lan and Joost Berkhout},
  title = {{PyJobShop}: Solving scheduling problems with constraint programming in {Python}},
  year = {2025},
  eprint = {2502.13483},
  eprinttype = {arxiv},
  note = {Preprint, version 1},
  doi = {10.48550/arXiv.2502.13483},
}

@article{Kasapidis2025,
  author = {Gregory A. Kasapidis and Dimitris C. Paraskevopoulos and Ioannis Mourtos and Panagiotis P. Repoussis},
  title = {A unified solution framework for flexible job shop scheduling problems with multiple resource constraints},
  year = {2025},
  journal = {European Journal of Operational Research},
  volume = {320},
  number = {3},
  pages = {479--495},
  doi = {10.1016/j.ejor.2024.08.010},
}

@article{Perrachon2025,
  author = {Quentin Perrachon and Alexandru-Liviu Olteanu and Marc Sevaux and Sylvain Fréchengues and Jean-François Kerviche},
  title = {Industrial multi-resource flexible job shop scheduling with partially necessary resources},
  year = {2025},
  journal = {European Journal of Operational Research},
  volume = {320},
  number = {2},
  pages = {309--327},
  doi = {10.1016/j.ejor.2024.07.023},
}

@article{Yasari2025,
  author = {Peyman Yasari and El-Houssaine Aghezzaf and Dieter Claeys},
  title = {Ergonomics-aware operator-task allocation and scheduling in flexible job shop systems},
  year = {2025},
  journal = {International Journal of Production Research},
  volume = {63},
  number = {18},
  pages = {6895--6914},
  doi = {10.1080/00207543.2025.2490826},
}

@article{Terbrack2025,
  author = {Hajo Terbrack and Thorsten Claus},
  title = {The generalized energy-aware flexible job shop scheduling model: A constraint programming approach},
  year = {2025},
  journal = {Computers \& Industrial Engineering},
  volume = {204},
  eid = {111065},
  doi = {10.1016/j.cie.2025.111065},
}

@article{Ferreira2026,
  author = {Ferreira, Guilherme de Souza and Mateus, Geraldo Robson and Ravetti, Martín Gómez},
  title = {Mathematical formulations and an enhanced column generation for the hybrid flow shop scheduling problem},
  year = {2026},
  journal = {Optimization Letters},
  doi = {10.1007/s11590-026-02280-2},
}

@article{Chen2025,
  author = {Xiaolong Chen and Junqing Li and Zunxun Wang and Qingda Chen and Kaizhou Gao and Quanke Pan},
  title = {Optimizing Dynamic Flexible Job Shop Scheduling Using an Evolutionary Multitask Optimization Framework and Genetic Programming},
  year = {2025},
  journal = {IEEE Transactions on Evolutionary Computation},
  volume = {29},
  number = {5},
  pages = {1502--1516},
  doi = {10.1109/TEVC.2025.3543770},
}

@article{Zhang2026,
  author = {Rui Zhang and Jianwei Niu and Xuefeng Liu and Shaojie Tang and Jing Yuan},
  title = {Learning to Optimize Job Shop Scheduling Under Structural Uncertainty},
  year = {2026},
  journal = {Proceedings of the AAAI Conference on Artificial Intelligence},
  volume = {40},
  number = {43},
  pages = {36509--36517},
  doi = {10.1609/aaai.v40i43.40973},
}

@article{DaCol2022,
  author = {Da Col, Giacomo and Teppan, Erich C.},
  title = {Industrial-size job shop scheduling with constraint programming},
  year = {2022},
  journal = {Operations Research Perspectives},
  volume = {9},
  eid = {100249},
  doi = {10.1016/j.orp.2022.100249},
}

% The benchmark appendix begins with full-width, multipage tables.
\onecolumn
\appendix
\section{Supplementary benchmark detail}\label{sec:appendix}

\subsection{Problem-class summaries and aggregate weighting}

The summaries below use the same 69 instances and seed-median aggregation
as the main text. Table~\ref{tab:classes} gives all quality and time measures
by problem class. Table~\ref{tab:class-balanced} first averages within each
class and then weights the three class means equally, providing a view that
is less influenced by their different sample sizes.

{\small\setlength{\tabcolsep}{6pt}
\begin{longtable}{llrrrr}
\caption{Class summaries: means of instance-level seed medians. Lower is better. Bold marks minima within each class; XG is excluded from the runtime minimum.}\label{tab:classes}\\
\toprule
Class & Method & HV (\%) & MS (\%) & FT (\%) & Time (s) \\
\midrule
\endfirsthead
\multicolumn{6}{l}{\small\itshape Table \thetable\ continued}\\
\toprule
Class & Method & HV (\%) & MS (\%) & FT (\%) & Time (s) \\
\midrule
\endhead
\midrule\multicolumn{6}{r}{\small Continued on next page}\\
\endfoot
\bottomrule
\endlastfoot
JSP & XG & 3.560 & 16.837 & 11.323 & 0.022 \\
JSP & XH & 0.271 & 1.366 & 1.729 & 26.454 \\
JSP & XT & 0.434 & 3.013 & 3.335 & 22.143 \\
JSP & XE & 2.309 & 12.461 & 9.965 & 5.580 \\
JSP & XHT & \textbf{0.192} & 1.193 & \textbf{1.145} & 27.275 \\
JSP & XHE & 0.331 & 1.516 & 2.116 & 16.561 \\
JSP & XTE & 0.490 & 3.279 & 3.736 & 14.613 \\
JSP & XHTE & 0.224 & \textbf{1.126} & 1.946 & 19.498 \\
JSP & NSGA-II & 1.620 & 7.030 & 13.245 & \textbf{1.019} \\
JSP & SPEA2 & 1.567 & 7.174 & 12.745 & 1.046 \\
JSP & MOEA/D & 1.575 & 6.619 & 12.820 & 1.201 \\
JSP & SMS-EMOA & 1.606 & 7.160 & 13.029 & 1.021 \\
FJSP & XG & 2.849 & 21.472 & 15.227 & 0.027 \\
FJSP & XH & \textbf{0.171} & \textbf{0.907} & \textbf{1.829} & 31.733 \\
FJSP & XT & 1.108 & 7.943 & 6.908 & 27.442 \\
FJSP & XE & 2.102 & 15.841 & 12.147 & 3.452 \\
FJSP & XHT & 0.193 & 0.921 & 2.264 & 30.561 \\
FJSP & XHE & 0.270 & 1.403 & 2.688 & 17.496 \\
FJSP & XTE & 1.100 & 7.943 & 6.975 & 15.574 \\
FJSP & XHTE & 0.238 & 1.134 & 2.881 & 21.775 \\
FJSP & NSGA-II & 4.663 & 44.322 & 47.019 & 0.359 \\
FJSP & SPEA2 & 4.545 & 43.172 & 45.752 & 0.387 \\
FJSP & MOEA/D & 3.757 & 36.495 & 40.401 & 0.528 \\
FJSP & SMS-EMOA & 4.688 & 44.140 & 46.873 & \textbf{0.357} \\
PFSP & XG & 4.589 & 14.609 & 16.895 & 0.064 \\
PFSP & XH & 2.853 & 8.540 & 12.136 & 68.303 \\
PFSP & XT & 2.799 & 9.188 & 10.940 & 64.331 \\
PFSP & XE & 0.343 & 1.718 & 1.776 & 63.810 \\
PFSP & XHT & 2.247 & 6.827 & 10.104 & 69.519 \\
PFSP & XHE & 0.958 & 2.976 & 4.693 & 66.675 \\
PFSP & XTE & 0.898 & 3.183 & 4.148 & 65.144 \\
PFSP & XHTE & 1.091 & 3.376 & 5.586 & 67.703 \\
PFSP & NSGA-II & 1.064 & 3.832 & 3.910 & \textbf{0.363} \\
PFSP & SPEA2 & 1.002 & 3.576 & 3.973 & 0.384 \\
PFSP & MOEA/D & \textbf{0.276} & \textbf{0.957} & \textbf{1.529} & 0.535 \\
PFSP & SMS-EMOA & 1.094 & 3.839 & 4.267 & 0.368 \\
\end{longtable}
}}

\begin{table}[htbp]
\centering\small\setlength{\tabcolsep}{5pt}
\caption{Supplementary overall summary with equal weights for JSP, FJSP
and PFSP. Each entry is the mean of the three class-level values in
Table~\ref{tab:classes}. Lower is better; bold marks column minima,
excluding the one-construction XG reference from the runtime minimum.}
\label{tab:class-balanced}
\begin{tabular}{lrrrr}
\toprule
Method & HV deficit (\%) & MS gap (\%) & FT gap (\%) & Time (s) \\
\midrule
XG & 3.666 & 17.639 & 14.482 & 0.037 \\
XH & 1.098 & 3.604 & 5.231 & 42.164 \\
XT & 1.447 & 6.714 & 7.061 & 37.972 \\
XE & 1.585 & 10.007 & 7.963 & 24.281 \\
XHT & 0.877 & 2.981 & 4.504 & 42.452 \\
XHE & 0.520 & 1.965 & \textbf{3.165} & 33.577 \\
XTE & 0.829 & 4.802 & 4.953 & 31.777 \\
XHTE & \textbf{0.518} & \textbf{1.879} & 3.471 & 36.325 \\
\midrule
NSGA-II & 2.449 & 18.395 & 21.391 & \textbf{0.580} \\
SPEA2 & 2.371 & 17.974 & 20.824 & 0.606 \\
MOEA/D & 1.869 & 14.690 & 18.250 & 0.755 \\
SMS-EMOA & 2.463 & 18.380 & 21.389 & 0.582 \\
\bottomrule
\end{tabular}
}
\end{table}
\FloatBarrier

\subsection{Separate objective distributions}

\begin{figure}[!htbp]
\centering\includegraphics[width=\textwidth]{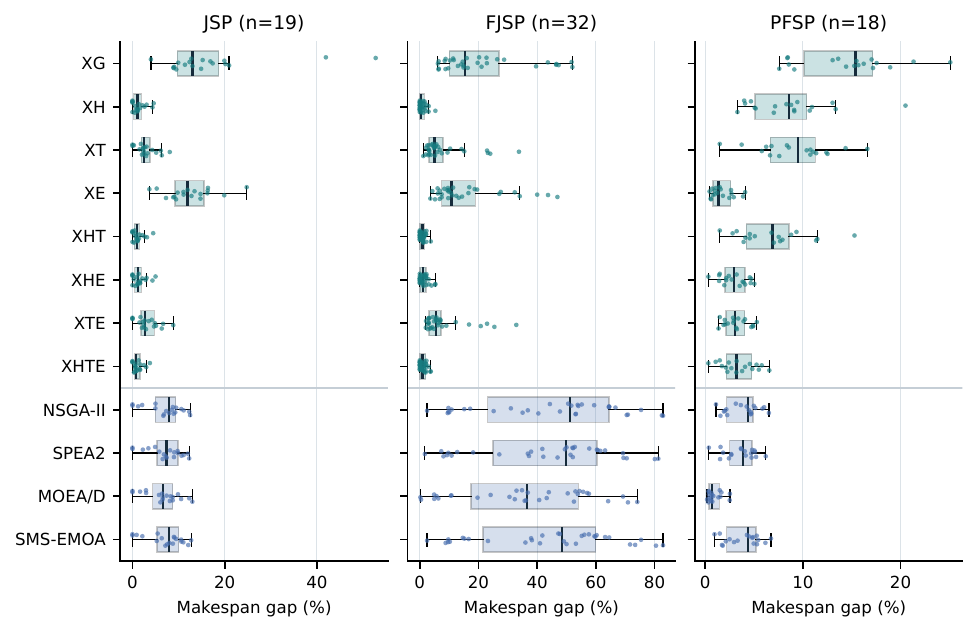}
\caption{Makespan gaps relative to the lowest observed makespan on each
instance. Each point is one median over three seeds; boxes and whiskers
follow the convention in Figure~\ref{fig:quality}. All observations are
retained, and horizontal scales differ between panels.}
\label{fig:makespan}
\end{figure}

\begin{figure}[!htbp]
\centering\includegraphics[width=\textwidth]{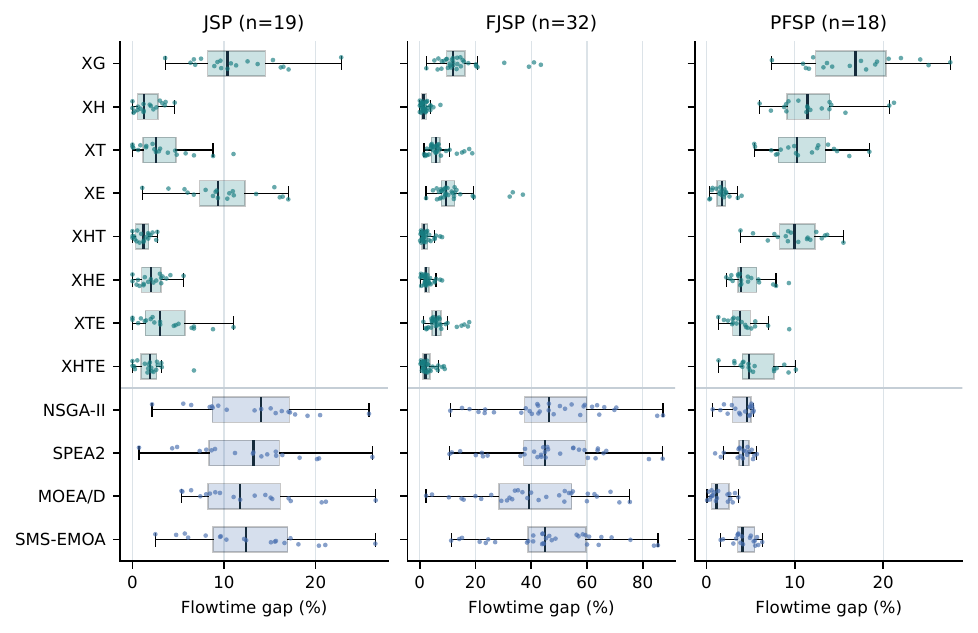}
\caption{Total job-flowtime gaps relative to the lowest observed value
on each instance, with the same aggregation and display conventions as
Figure~\ref{fig:makespan}. A method's best makespan and best flowtime
need not occur in the same schedule.}
\label{fig:flowtime}
\end{figure}
\FloatBarrier

\subsection{Instance inventory and per-instance results}

Table~\ref{tab:instances} defines the manuscript's comparison set in full.
Processing times retain their source units. A machine count in an FJSP
instance includes the available alternatives, not the number visited by
every job.

Tables~\ref{tab:instance-quality} and~\ref{tab:instance-time} report one
median over three seeds for each instance and method. Bold entries use
unrounded values, so identically rounded cells can differ in emphasis.
Individual seeds, MS/FT values, archives and evaluation traces are retained
in the accompanying machine-readable results.

{\footnotesize\setlength{\tabcolsep}{5pt}\renewcommand{\arraystretch}{0.97}
\begin{longtable}{lllrrr}
\caption{The 69 instances included in the manuscript. Source identifiers are retained; Taillard JSP and PFSP use separate identifier prefixes.}\label{tab:instances}\\
\toprule
Class & Instance & Family & Jobs & Machines & Operations \\
\midrule
\endfirsthead
\multicolumn{6}{l}{\small\itshape Table \thetable\ continued}\\
\toprule
Class & Instance & Family & Jobs & Machines & Operations \\
\midrule
\endhead
\midrule\multicolumn{6}{r}{\small Continued on next page}\\
\endfoot
\bottomrule
\endlastfoot
JSP & abz5 & OR-Library & 10 & 10 & 100 \\
JSP & ft06 & OR-Library & 6 & 6 & 36 \\
JSP & la01 & OR-Library & 10 & 5 & 50 \\
JSP & la06 & OR-Library & 15 & 5 & 75 \\
JSP & la16 & OR-Library & 10 & 10 & 100 \\
JSP & tai\_jsp002 & Taillard & 15 & 15 & 225 \\
JSP & tai\_jsp006 & Taillard & 15 & 15 & 225 \\
JSP & tai\_jsp013 & Taillard & 20 & 15 & 300 \\
JSP & tai\_jsp020 & Taillard & 20 & 15 & 300 \\
JSP & tai\_jsp025 & Taillard & 20 & 20 & 400 \\
JSP & tai\_jsp026 & Taillard & 20 & 20 & 400 \\
JSP & tai\_jsp031 & Taillard & 30 & 15 & 450 \\
JSP & tai\_jsp037 & Taillard & 30 & 15 & 450 \\
JSP & tai\_jsp044 & Taillard & 30 & 20 & 600 \\
JSP & tai\_jsp049 & Taillard & 30 & 20 & 600 \\
JSP & tai\_jsp052 & Taillard & 50 & 15 & 750 \\
JSP & tai\_jsp053 & Taillard & 50 & 15 & 750 \\
JSP & tai\_jsp068 & Taillard & 50 & 20 & 1000 \\
JSP & tai\_jsp070 & Taillard & 50 & 20 & 1000 \\
FJSP & lar01\_1 & Behnke-Geiger & 10 & 60 & 50 \\
FJSP & lar01\_3 & Behnke-Geiger & 10 & 60 & 50 \\
FJSP & lar02\_2 & Behnke-Geiger & 20 & 60 & 100 \\
FJSP & lar02\_3 & Behnke-Geiger & 20 & 60 & 100 \\
FJSP & lar03\_4 & Behnke-Geiger & 50 & 60 & 250 \\
FJSP & med01\_2 & Behnke-Geiger & 10 & 40 & 50 \\
FJSP & med01\_3 & Behnke-Geiger & 10 & 40 & 50 \\
FJSP & med02\_3 & Behnke-Geiger & 20 & 40 & 100 \\
FJSP & med02\_5 & Behnke-Geiger & 20 & 40 & 100 \\
FJSP & med03\_1 & Behnke-Geiger & 50 & 40 & 250 \\
FJSP & med03\_2 & Behnke-Geiger & 50 & 40 & 250 \\
FJSP & mk01 & Brandimarte & 10 & 6 & 55 \\
FJSP & mk02 & Brandimarte & 10 & 6 & 58 \\
FJSP & mk03 & Brandimarte & 15 & 8 & 150 \\
FJSP & mk04 & Brandimarte & 15 & 8 & 90 \\
FJSP & mk05 & Brandimarte & 15 & 4 & 106 \\
FJSP & mk06 & Brandimarte & 10 & 10 & 150 \\
FJSP & mk07 & Brandimarte & 20 & 5 & 100 \\
FJSP & mk08 & Brandimarte & 20 & 10 & 225 \\
FJSP & mk09 & Brandimarte & 20 & 10 & 240 \\
FJSP & mk10 & Brandimarte & 20 & 15 & 240 \\
FJSP & mk11 & Brandimarte & 30 & 5 & 179 \\
FJSP & mk12 & Brandimarte & 30 & 10 & 193 \\
FJSP & mk13 & Brandimarte & 30 & 10 & 231 \\
FJSP & mk14 & Brandimarte & 30 & 15 & 277 \\
FJSP & mk15 & Brandimarte & 30 & 15 & 284 \\
FJSP & sm01\_1 & Behnke-Geiger & 10 & 20 & 50 \\
FJSP & sm01\_2 & Behnke-Geiger & 10 & 20 & 50 \\
FJSP & sm02\_4 & Behnke-Geiger & 20 & 20 & 100 \\
FJSP & sm02\_5 & Behnke-Geiger & 20 & 20 & 100 \\
FJSP & sm03\_3 & Behnke-Geiger & 50 & 20 & 250 \\
FJSP & sm03\_4 & Behnke-Geiger & 50 & 20 & 250 \\
PFSP & ta001 & Taillard & 20 & 5 & 100 \\
PFSP & ta002 & Taillard & 20 & 5 & 100 \\
PFSP & ta003 & Taillard & 20 & 5 & 100 \\
PFSP & ta004 & Taillard & 20 & 5 & 100 \\
PFSP & ta005 & Taillard & 20 & 5 & 100 \\
PFSP & ta008 & Taillard & 20 & 5 & 100 \\
PFSP & ta013 & Taillard & 20 & 10 & 200 \\
PFSP & ta018 & Taillard & 20 & 10 & 200 \\
PFSP & ta021 & Taillard & 20 & 20 & 400 \\
PFSP & ta025 & Taillard & 20 & 20 & 400 \\
PFSP & ta038 & Taillard & 50 & 5 & 250 \\
PFSP & ta039 & Taillard & 50 & 5 & 250 \\
PFSP & ta041 & Taillard & 50 & 10 & 500 \\
PFSP & ta049 & Taillard & 50 & 10 & 500 \\
PFSP & ta059 & Taillard & 50 & 20 & 1000 \\
PFSP & ta060 & Taillard & 50 & 20 & 1000 \\
PFSP & ta064 & Taillard & 100 & 5 & 500 \\
PFSP & ta069 & Taillard & 100 & 5 & 500 \\
\end{longtable}
}}

\clearpage
{\footnotesize\setlength{\tabcolsep}{2pt}\renewcommand{\arraystretch}{1.0}
\begin{longtable}{llrrrrrrrrrrrr}
\caption{Per-instance HV deficit (percent): median over three seeds. Bold marks the lowest unrounded value within each instance; ties are retained.}\label{tab:instance-quality}\\
\toprule
Class & Instance & XG & XH & XT & XE & XHT & XHE & XTE & XHTE & NSGA-II & SPEA2 & MOEA/D & SMS-EMOA \\
\midrule
\endfirsthead
\multicolumn{14}{l}{\small\itshape Table \thetable\ continued}\\
\toprule
Class & Instance & XG & XH & XT & XE & XHT & XHE & XTE & XHTE & NSGA-II & SPEA2 & MOEA/D & SMS-EMOA \\
\midrule
\endhead
\midrule\multicolumn{14}{r}{\small Continued on next page}\\
\endfoot
\bottomrule
\endlastfoot
JSP & abz5 & 10.320 & 0.431 & 0.669 & 6.432 & \textbf{0.157} & 0.498 & 0.643 & 0.306 & 1.694 & 1.503 & 1.421 & 1.746 \\
JSP & ft06 & 18.652 & 0.206 & \textbf{0.008} & 5.782 & 0.200 & 0.195 & 0.044 & 0.208 & 0.557 & 0.194 & 1.388 & 1.096 \\
JSP & la01 & 4.817 & 0.687 & \textbf{0.243} & 2.153 & 0.498 & 0.785 & 0.252 & 0.489 & 1.807 & 1.690 & 1.117 & 1.168 \\
JSP & la06 & 2.187 & 0.780 & \textbf{0.249} & 1.930 & 0.367 & 0.942 & 0.326 & 0.315 & 1.607 & 1.553 & 1.533 & 1.356 \\
JSP & la16 & 4.661 & 0.719 & \textbf{0.000} & 2.760 & 0.634 & 0.869 & 0.219 & 0.217 & 0.887 & 0.987 & 1.396 & 1.150 \\
JSP & tai\_jsp002 & 3.181 & 0.140 & 0.327 & 2.707 & \textbf{0.025} & 0.103 & 0.327 & \textbf{0.025} & 1.726 & 1.566 & 1.617 & 1.840 \\
JSP & tai\_jsp006 & 3.072 & 0.369 & 0.140 & 2.510 & \textbf{0.120} & 0.492 & 0.366 & 0.246 & 1.825 & 1.768 & 1.866 & 1.440 \\
JSP & tai\_jsp013 & 2.043 & 0.545 & 0.480 & 1.878 & \textbf{0.380} & 0.545 & 0.481 & 0.446 & 1.743 & 1.874 & 1.775 & 2.019 \\
JSP & tai\_jsp020 & 1.511 & 0.221 & 0.352 & 1.471 & \textbf{0.053} & 0.194 & 0.352 & \textbf{0.053} & 2.207 & 1.999 & 1.848 & 1.930 \\
JSP & tai\_jsp025 & 2.348 & 0.263 & 0.410 & 2.243 & 0.264 & 0.368 & 0.460 & \textbf{0.216} & 1.539 & 1.757 & 1.289 & 1.652 \\
JSP & tai\_jsp026 & 2.463 & 0.205 & 0.588 & 1.973 & \textbf{0.119} & 0.204 & 0.694 & 0.234 & 1.737 & 1.736 & 1.413 & 1.820 \\
JSP & tai\_jsp031 & 1.464 & 0.144 & 0.410 & 1.428 & \textbf{0.096} & 0.144 & 0.410 & 0.214 & 1.728 & 1.616 & 1.601 & 1.528 \\
JSP & tai\_jsp037 & 2.426 & \textbf{0.111} & 0.837 & 2.306 & 0.155 & 0.173 & 0.953 & 0.173 & 2.042 & 1.871 & 2.115 & 1.848 \\
JSP & tai\_jsp044 & 1.400 & 0.087 & 0.556 & 1.381 & 0.078 & 0.160 & 0.560 & \textbf{0.056} & 1.626 & 1.634 & 1.551 & 1.629 \\
JSP & tai\_jsp049 & 1.616 & \textbf{0.059} & 0.544 & 1.612 & 0.224 & 0.256 & 0.703 & 0.254 & 1.484 & 1.313 & 1.173 & 1.558 \\
JSP & tai\_jsp052 & 1.522 & \textbf{0.029} & 0.744 & 1.514 & 0.080 & 0.080 & 0.744 & 0.383 & 2.078 & 2.090 & 2.159 & 2.084 \\
JSP & tai\_jsp053 & 1.211 & 0.080 & 0.417 & 1.074 & \textbf{0.065} & 0.091 & 0.510 & 0.185 & 1.513 & 1.590 & 1.594 & 1.606 \\
JSP & tai\_jsp068 & 1.326 & \textbf{0.058} & 0.673 & 1.313 & 0.112 & 0.112 & 0.673 & 0.163 & 1.649 & 1.647 & 1.695 & 1.654 \\
JSP & tai\_jsp070 & 1.413 & 0.012 & 0.593 & 1.407 & \textbf{0.012} & 0.078 & 0.591 & 0.075 & 1.334 & 1.387 & 1.372 & 1.398 \\
FJSP & lar01\_1 & 1.919 & \textbf{0.000} & 0.480 & 1.618 & 0.020 & 0.020 & 0.506 & 0.020 & 5.174 & 5.374 & 3.803 & 4.858 \\
FJSP & lar01\_3 & 2.006 & \textbf{0.197} & 0.599 & 1.129 & 0.266 & 0.266 & 0.674 & 0.293 & 4.646 & 5.081 & 3.595 & 4.933 \\
FJSP & lar02\_2 & 0.982 & 0.034 & 0.319 & 0.658 & 0.034 & 0.067 & 0.285 & \textbf{0.034} & 3.834 & 3.684 & 3.923 & 4.052 \\
FJSP & lar02\_3 & 0.713 & \textbf{0.074} & 0.294 & 0.461 & \textbf{0.074} & 0.153 & 0.295 & \textbf{0.074} & 4.187 & 3.928 & 4.069 & 4.200 \\
FJSP & lar03\_4 & 0.448 & 0.031 & 0.146 & 0.419 & \textbf{0.031} & 0.041 & 0.147 & 0.048 & 3.056 & 2.996 & 3.028 & 2.979 \\
FJSP & med01\_2 & 2.134 & \textbf{0.034} & 0.631 & 1.518 & \textbf{0.034} & 0.192 & 0.706 & 0.061 & 5.552 & 4.493 & 4.649 & 4.571 \\
FJSP & med01\_3 & 2.278 & 0.041 & 0.530 & 1.579 & \textbf{0.027} & 0.041 & 0.661 & \textbf{0.027} & 5.257 & 5.120 & 4.194 & 5.500 \\
FJSP & med02\_3 & 0.884 & 0.031 & 0.346 & 0.755 & 0.058 & 0.052 & 0.381 & \textbf{0.015} & 4.293 & 4.082 & 3.821 & 4.151 \\
FJSP & med02\_5 & 0.575 & \textbf{0.068} & 0.392 & 0.526 & 0.071 & 0.071 & 0.433 & 0.071 & 4.544 & 4.385 & 3.742 & 4.329 \\
FJSP & med03\_1 & 0.429 & \textbf{0.018} & 0.239 & 0.369 & 0.072 & 0.072 & 0.254 & 0.088 & 3.282 & 3.195 & 2.954 & 3.223 \\
FJSP & med03\_2 & 0.337 & 0.077 & 0.132 & 0.268 & 0.076 & 0.075 & 0.135 & \textbf{0.072} & 3.331 & 3.083 & 2.889 & 3.295 \\
FJSP & mk01 & 4.345 & 0.875 & 1.630 & 3.049 & 0.398 & 0.435 & 1.508 & \textbf{0.358} & 3.477 & 2.963 & 1.705 & 4.168 \\
FJSP & mk02 & 5.656 & 0.290 & 1.074 & 3.019 & \textbf{0.068} & 0.323 & 0.653 & 0.228 & 5.623 & 6.656 & 3.749 & 6.777 \\
FJSP & mk03 & 6.240 & \textbf{0.000} & 2.298 & 4.810 & 0.329 & 0.329 & 2.082 & 0.375 & 4.434 & 3.693 & 3.498 & 4.170 \\
FJSP & mk04 & 5.001 & 0.598 & 0.897 & 2.522 & 0.073 & 0.536 & 0.762 & \textbf{0.073} & 3.165 & 3.051 & 1.393 & 3.249 \\
FJSP & mk05 & 8.680 & \textbf{0.080} & 3.469 & 6.173 & 1.396 & 1.395 & 3.470 & 1.441 & 6.704 & 6.357 & 5.240 & 6.690 \\
FJSP & mk06 & 0.883 & 0.187 & 0.411 & 0.823 & 0.187 & \textbf{0.101} & 0.394 & 0.377 & 7.632 & 6.884 & 6.992 & 7.862 \\
FJSP & mk07 & 5.640 & 0.542 & 2.656 & 3.506 & \textbf{0.481} & 0.599 & 2.787 & 0.559 & 6.830 & 7.102 & 6.139 & 6.975 \\
FJSP & mk08 & 5.066 & 0.378 & 1.784 & 3.096 & 0.177 & 0.377 & 1.716 & \textbf{0.168} & 1.834 & 1.700 & 1.465 & 1.881 \\
FJSP & mk09 & 7.207 & \textbf{0.035} & 2.482 & 6.188 & 0.370 & 0.370 & 2.572 & 0.404 & 5.027 & 5.073 & 3.941 & 4.820 \\
FJSP & mk10 & 3.818 & 0.200 & 2.556 & 3.473 & 0.154 & 0.239 & 2.487 & \textbf{0.154} & 5.550 & 5.892 & 5.086 & 5.584 \\
FJSP & mk11 & 4.183 & 0.109 & 1.725 & 3.226 & \textbf{0.032} & 0.232 & 1.953 & 0.169 & 4.630 & 4.353 & 3.023 & 4.750 \\
FJSP & mk12 & 3.511 & \textbf{0.167} & 2.595 & 3.337 & 0.449 & 0.450 & 2.526 & 0.757 & 3.649 & 3.587 & 2.361 & 3.499 \\
FJSP & mk13 & 3.750 & \textbf{0.117} & 2.105 & 3.206 & 0.128 & 0.261 & 1.858 & 0.157 & 4.640 & 4.736 & 4.088 & 4.803 \\
FJSP & mk14 & 5.537 & 0.297 & 2.416 & 4.226 & \textbf{0.247} & 0.273 & 2.263 & 0.273 & 2.954 & 2.735 & 1.478 & 2.742 \\
FJSP & mk15 & 3.623 & \textbf{0.062} & 1.836 & 3.297 & 0.119 & 0.119 & 1.908 & 0.320 & 3.999 & 4.060 & 3.945 & 4.027 \\
FJSP & sm01\_1 & 1.597 & 0.450 & 0.295 & 0.903 & \textbf{0.077} & 0.450 & 0.498 & 0.084 & 5.679 & 5.698 & 3.940 & 5.962 \\
FJSP & sm01\_2 & 1.690 & \textbf{0.219} & 0.283 & 1.410 & 0.288 & 0.430 & 0.374 & 0.288 & 6.262 & 5.610 & 3.999 & 5.814 \\
FJSP & sm02\_4 & 0.505 & \textbf{0.178} & 0.217 & 0.492 & 0.191 & 0.294 & 0.219 & 0.179 & 5.423 & 5.221 & 4.965 & 5.312 \\
FJSP & sm02\_5 & 0.598 & \textbf{0.000} & 0.239 & 0.436 & 0.162 & 0.162 & 0.245 & 0.269 & 5.437 & 5.555 & 4.426 & 5.665 \\
FJSP & sm03\_3 & 0.567 & \textbf{0.042} & 0.325 & 0.478 & 0.044 & 0.173 & 0.314 & 0.056 & 4.538 & 4.595 & 4.151 & 4.640 \\
FJSP & sm03\_4 & 0.351 & \textbf{0.031} & 0.071 & 0.305 & \textbf{0.031} & \textbf{0.031} & 0.137 & 0.119 & 4.581 & 4.510 & 3.980 & 4.525 \\
PFSP & ta001 & 3.876 & 2.355 & 1.001 & 0.338 & 0.683 & 0.672 & 0.317 & 0.320 & 1.134 & 0.914 & \textbf{0.206} & 0.654 \\
PFSP & ta002 & 5.879 & 2.751 & 3.182 & 0.168 & 2.334 & 1.398 & 1.071 & 1.332 & 0.482 & 0.535 & \textbf{0.159} & 0.567 \\
PFSP & ta003 & 6.964 & 4.764 & 4.232 & 0.291 & 4.531 & 0.756 & 1.434 & 0.694 & 1.301 & 1.548 & \textbf{0.036} & 1.687 \\
PFSP & ta004 & 7.747 & 6.280 & 2.985 & \textbf{0.204} & 2.932 & 1.178 & 1.240 & 0.914 & 1.702 & 1.225 & 0.279 & 1.814 \\
PFSP & ta005 & 4.644 & 1.845 & 2.272 & 0.601 & 1.439 & 0.617 & 0.710 & 0.609 & 1.930 & 1.815 & \textbf{0.198} & 1.655 \\
PFSP & ta008 & 6.531 & 3.432 & 3.534 & \textbf{0.429} & 2.242 & 1.464 & 0.865 & 1.612 & 2.248 & 1.757 & 0.961 & 2.109 \\
PFSP & ta013 & 5.531 & 2.845 & 3.662 & \textbf{0.210} & 2.719 & 0.818 & 1.067 & 1.771 & 1.241 & 1.069 & 0.355 & 1.289 \\
PFSP & ta018 & 7.118 & 5.642 & 4.338 & 0.935 & 3.894 & 1.181 & 1.378 & 1.398 & 1.289 & 1.398 & \textbf{0.239} & 1.505 \\
PFSP & ta021 & 3.108 & 1.671 & 1.851 & 0.103 & 1.344 & 0.666 & 0.401 & 0.515 & 0.214 & 0.232 & \textbf{0.035} & 0.307 \\
PFSP & ta025 & 2.804 & 2.001 & 1.828 & \textbf{0.309} & 1.557 & 0.550 & 0.527 & 0.792 & 0.669 & 0.582 & 0.331 & 0.738 \\
PFSP & ta038 & 4.173 & 2.410 & 2.981 & \textbf{0.275} & 2.173 & 0.491 & 0.841 & 0.552 & 0.923 & 0.981 & 0.550 & 1.329 \\
PFSP & ta039 & 7.005 & 2.944 & 5.061 & 0.477 & 2.790 & 1.744 & 0.779 & 2.247 & 1.034 & 1.251 & \textbf{0.253} & 1.208 \\
PFSP & ta041 & 3.148 & 2.691 & 2.475 & 0.294 & 2.362 & 0.837 & 1.090 & 1.258 & 0.934 & 0.875 & \textbf{0.194} & 0.688 \\
PFSP & ta049 & 2.938 & 2.053 & 2.229 & 0.711 & 1.907 & 1.274 & 1.037 & 1.444 & 1.129 & 1.079 & \textbf{0.162} & 1.125 \\
PFSP & ta059 & 2.206 & 1.242 & 1.446 & 0.333 & 1.273 & 0.608 & 0.549 & 0.781 & 0.502 & 0.501 & \textbf{0.199} & 0.490 \\
PFSP & ta060 & 1.891 & 1.137 & 1.198 & 0.325 & 0.990 & 0.658 & 0.438 & 0.672 & 0.549 & 0.457 & \textbf{0.104} & 0.541 \\
PFSP & ta064 & 2.918 & 1.922 & 2.560 & \textbf{0.057} & 1.940 & 0.716 & 0.747 & 0.983 & 0.997 & 0.949 & 0.312 & 0.917 \\
PFSP & ta069 & 4.114 & 3.366 & 3.543 & \textbf{0.107} & 3.339 & 1.615 & 1.675 & 1.739 & 0.871 & 0.866 & 0.389 & 1.065 \\
\end{longtable}
}}
\clearpage
{\footnotesize\setlength{\tabcolsep}{2pt}\renewcommand{\arraystretch}{1.0}
\begin{longtable}{llrrrrrrrrrrrr}
\caption{Per-instance internal wall time (seconds): median over three seeds. Bold marks the lowest unrounded time among the search methods. XG uses one construction.}\label{tab:instance-time}\\
\toprule
Class & Instance & XG & XH & XT & XE & XHT & XHE & XTE & XHTE & NSGA-II & SPEA2 & MOEA/D & SMS-EMOA \\
\midrule
\endfirsthead
\multicolumn{14}{l}{\small\itshape Table \thetable\ continued}\\
\toprule
Class & Instance & XG & XH & XT & XE & XHT & XHE & XTE & XHTE & NSGA-II & SPEA2 & MOEA/D & SMS-EMOA \\
\midrule
\endhead
\midrule\multicolumn{14}{r}{\small Continued on next page}\\
\endfoot
\bottomrule
\endlastfoot
JSP & abz5 & 0.003 & 2.644 & 2.117 & 1.186 & 2.459 & 1.838 & 1.648 & 2.030 & 0.193 & 0.222 & 0.329 & \textbf{0.188} \\
JSP & ft06 & 0.001 & 0.701 & 0.585 & 0.430 & 0.727 & 0.534 & 0.543 & 0.637 & \textbf{0.090} & 0.123 & 0.235 & 0.097 \\
JSP & la01 & 0.001 & 1.245 & 0.967 & 0.575 & 1.113 & 0.864 & 0.820 & 1.010 & \textbf{0.116} & 0.142 & 0.261 & 0.121 \\
JSP & la06 & 0.002 & 2.154 & 1.889 & 0.914 & 1.968 & 1.542 & 1.361 & 1.699 & 0.168 & 0.188 & 0.327 & \textbf{0.160} \\
JSP & la16 & 0.002 & 2.312 & 1.926 & 1.103 & 2.379 & 1.891 & 1.615 & 1.938 & \textbf{0.190} & 0.219 & 0.380 & 0.208 \\
JSP & tai\_jsp002 & 0.007 & 7.756 & 6.485 & 2.696 & 7.614 & 5.159 & 4.619 & 5.985 & \textbf{0.378} & 0.435 & 0.601 & 0.385 \\
JSP & tai\_jsp006 & 0.007 & 7.883 & 6.775 & 2.728 & 7.748 & 5.309 & 4.878 & 5.974 & \textbf{0.409} & 0.431 & 0.564 & 0.412 \\
JSP & tai\_jsp013 & 0.010 & 12.113 & 9.923 & 3.760 & 12.260 & 7.676 & 6.995 & 9.093 & 0.589 & \textbf{0.568} & 0.796 & 0.587 \\
JSP & tai\_jsp020 & 0.010 & 12.021 & 9.985 & 3.598 & 11.434 & 7.645 & 7.351 & 9.351 & \textbf{0.555} & 0.610 & 0.761 & 0.562 \\
JSP & tai\_jsp025 & 0.015 & 17.500 & 14.166 & 5.230 & 16.096 & 11.409 & 9.463 & 12.076 & \textbf{0.770} & 0.813 & 0.916 & 0.792 \\
JSP & tai\_jsp026 & 0.013 & 17.929 & 14.011 & 5.285 & 17.655 & 11.625 & 10.540 & 12.992 & 0.780 & \textbf{0.756} & 0.890 & 0.792 \\
JSP & tai\_jsp031 & 0.021 & 26.020 & 20.587 & 5.687 & 23.239 & 16.759 & 14.267 & 17.508 & \textbf{0.933} & 1.016 & 1.164 & 0.978 \\
JSP & tai\_jsp037 & 0.022 & 24.858 & 22.346 & 6.293 & 27.760 & 15.115 & 14.138 & 18.730 & \textbf{0.945} & 1.032 & 1.239 & 1.005 \\
JSP & tai\_jsp044 & 0.029 & 36.945 & 27.509 & 8.138 & 35.177 & 21.979 & 18.532 & 26.007 & 1.315 & 1.325 & 1.418 & \textbf{1.272} \\
JSP & tai\_jsp049 & 0.028 & 33.987 & 28.221 & 8.313 & 35.449 & 20.420 & 18.563 & 25.560 & \textbf{1.269} & 1.323 & 1.456 & 1.383 \\
JSP & tai\_jsp052 & 0.049 & 62.492 & 57.313 & 10.026 & 65.441 & 36.669 & 30.281 & 46.494 & 2.143 & 2.139 & 2.278 & \textbf{2.053} \\
JSP & tai\_jsp053 & 0.051 & 62.994 & 52.147 & 11.288 & 65.797 & 39.293 & 32.299 & 48.077 & 2.271 & 2.253 & 2.481 & \textbf{2.205} \\
JSP & tai\_jsp068 & 0.068 & 87.391 & 71.611 & 15.192 & 92.480 & 54.556 & 51.630 & 66.258 & 3.119 & 3.157 & 3.455 & \textbf{3.076} \\
JSP & tai\_jsp070 & 0.072 & 83.688 & 72.152 & 13.585 & 91.420 & 54.371 & 48.111 & 59.053 & 3.128 & \textbf{3.122} & 3.275 & 3.127 \\
FJSP & lar01\_1 & 0.018 & 14.544 & 15.162 & 2.486 & 11.510 & 7.861 & 8.935 & 8.873 & 0.149 & 0.174 & 0.320 & \textbf{0.137} \\
FJSP & lar01\_3 & 0.017 & 12.940 & 12.968 & 2.269 & 8.257 & 6.163 & 8.002 & 6.124 & 0.161 & 0.164 & 0.302 & \textbf{0.137} \\
FJSP & lar02\_2 & 0.043 & 42.268 & 38.028 & 4.512 & 44.335 & 23.283 & 20.869 & 30.488 & 0.248 & 0.248 & 0.381 & \textbf{0.222} \\
FJSP & lar02\_3 & 0.036 & 41.515 & 38.607 & 4.190 & 36.731 & 21.405 & 20.254 & 26.203 & \textbf{0.218} & 0.268 & 0.414 & 0.235 \\
FJSP & lar03\_4 & 0.135 & 121.307 & 148.238 & 13.079 & 121.720 & 73.123 & 78.735 & 100.396 & \textbf{0.557} & 0.603 & 0.691 & 0.590 \\
FJSP & med01\_2 & 0.012 & 8.163 & 8.744 & 1.826 & 8.108 & 5.101 & 5.542 & 5.950 & 0.146 & 0.160 & 0.307 & \textbf{0.145} \\
FJSP & med01\_3 & 0.011 & 9.856 & 9.113 & 1.843 & 9.690 & 5.586 & 5.353 & 7.066 & 0.156 & 0.150 & 0.315 & \textbf{0.140} \\
FJSP & med02\_3 & 0.034 & 34.561 & 28.851 & 3.289 & 29.841 & 18.852 & 17.921 & 23.367 & \textbf{0.215} & 0.241 & 0.356 & 0.224 \\
FJSP & med02\_5 & 0.031 & 32.750 & 26.545 & 3.358 & 32.802 & 17.879 & 16.188 & 21.837 & \textbf{0.226} & 0.246 & 0.367 & 0.238 \\
FJSP & med03\_1 & 0.101 & 131.180 & 115.426 & 8.729 & 129.460 & 70.469 & 54.767 & 84.209 & \textbf{0.530} & 0.581 & 0.730 & 0.533 \\
FJSP & med03\_2 & 0.097 & 131.619 & 101.174 & 9.219 & 119.687 & 65.930 & 61.018 & 82.789 & 0.579 & 0.587 & 0.676 & \textbf{0.512} \\
FJSP & mk01 & 0.002 & 2.392 & 1.722 & 0.751 & 2.176 & 1.464 & 1.188 & 1.688 & 0.142 & 0.176 & 0.299 & \textbf{0.134} \\
FJSP & mk02 & 0.003 & 3.966 & 2.427 & 0.907 & 3.836 & 2.280 & 1.848 & 2.931 & 0.144 & 0.173 & 0.318 & \textbf{0.139} \\
FJSP & mk03 & 0.007 & 9.816 & 7.405 & 2.169 & 9.936 & 6.160 & 4.632 & 7.296 & \textbf{0.362} & 0.400 & 0.494 & 0.366 \\
FJSP & mk04 & 0.004 & 4.114 & 3.694 & 1.185 & 4.050 & 2.727 & 2.445 & 3.067 & 0.240 & 0.240 & 0.419 & \textbf{0.234} \\
FJSP & mk05 & 0.004 & 4.951 & 3.279 & 1.392 & 4.842 & 3.097 & 2.406 & 3.451 & 0.270 & 0.340 & 0.470 & \textbf{0.268} \\
FJSP & mk06 & 0.008 & 9.612 & 7.918 & 2.397 & 8.976 & 5.398 & 5.307 & 6.450 & \textbf{0.328} & 0.369 & 0.511 & 0.343 \\
FJSP & mk07 & 0.006 & 8.676 & 5.608 & 1.516 & 8.698 & 4.564 & 3.655 & 5.920 & 0.263 & 0.294 & 0.453 & \textbf{0.262} \\
FJSP & mk08 & 0.009 & 10.884 & 8.697 & 2.959 & 11.030 & 6.986 & 5.882 & 8.313 & \textbf{0.554} & 0.618 & 0.820 & 0.611 \\
FJSP & mk09 & 0.012 & 19.885 & 12.615 & 3.596 & 18.073 & 10.245 & 7.688 & 13.440 & 0.624 & 0.625 & 0.853 & \textbf{0.580} \\
FJSP & mk10 & 0.015 & 26.987 & 14.811 & 3.978 & 23.702 & 14.800 & 9.742 & 16.477 & \textbf{0.555} & 0.649 & 0.759 & 0.573 \\
FJSP & mk11 & 0.010 & 13.963 & 10.541 & 2.450 & 13.775 & 8.225 & 6.458 & 10.075 & 0.525 & \textbf{0.523} & 0.684 & 0.531 \\
FJSP & mk12 & 0.010 & 13.183 & 10.403 & 2.367 & 12.107 & 7.812 & 6.853 & 9.637 & 0.515 & \textbf{0.507} & 0.644 & 0.517 \\
FJSP & mk13 & 0.021 & 37.671 & 23.673 & 3.860 & 37.207 & 19.611 & 13.128 & 25.171 & 0.555 & 0.629 & 0.820 & \textbf{0.554} \\
FJSP & mk14 & 0.016 & 21.374 & 18.233 & 3.761 & 20.412 & 13.123 & 10.702 & 14.388 & 0.686 & 0.757 & 0.801 & \textbf{0.684} \\
FJSP & mk15 & 0.023 & 37.053 & 24.114 & 4.618 & 36.677 & 23.015 & 14.962 & 28.301 & \textbf{0.656} & 0.738 & 0.882 & 0.656 \\
FJSP & sm01\_1 & 0.006 & 5.330 & 4.319 & 1.090 & 5.052 & 2.977 & 2.578 & 3.638 & 0.136 & 0.156 & 0.271 & \textbf{0.135} \\
FJSP & sm01\_2 & 0.007 & 5.862 & 4.666 & 1.066 & 5.576 & 3.356 & 2.957 & 3.764 & 0.137 & 0.155 & 0.315 & \textbf{0.118} \\
FJSP & sm02\_4 & 0.017 & 17.200 & 16.193 & 2.075 & 18.229 & 9.519 & 8.750 & 11.976 & \textbf{0.220} & 0.253 & 0.371 & 0.220 \\
FJSP & sm02\_5 & 0.014 & 17.842 & 14.562 & 1.977 & 16.796 & 9.247 & 8.357 & 11.192 & 0.227 & 0.242 & 0.413 & \textbf{0.227} \\
FJSP & sm03\_3 & 0.068 & 85.832 & 72.403 & 5.786 & 90.703 & 47.707 & 42.465 & 58.160 & 0.596 & 0.570 & 0.696 & \textbf{0.569} \\
FJSP & sm03\_4 & 0.065 & 78.175 & 68.020 & 5.766 & 73.972 & 41.910 & 38.792 & 54.154 & 0.578 & \textbf{0.564} & 0.746 & 0.584 \\
PFSP & ta001 & 0.010 & 9.833 & 9.375 & 9.467 & 9.440 & 9.266 & 9.183 & 9.413 & 0.135 & 0.153 & 0.278 & \textbf{0.123} \\
PFSP & ta002 & 0.011 & 10.005 & 9.785 & 9.588 & 10.759 & 9.639 & 9.891 & 9.415 & \textbf{0.128} & 0.149 & 0.286 & 0.135 \\
PFSP & ta003 & 0.012 & 9.983 & 10.599 & 9.430 & 10.044 & 9.806 & 9.420 & 9.764 & \textbf{0.125} & 0.158 & 0.263 & 0.128 \\
PFSP & ta004 & 0.011 & 9.997 & 9.789 & 9.783 & 10.119 & 8.970 & 9.280 & 9.511 & 0.135 & 0.150 & 0.287 & \textbf{0.127} \\
PFSP & ta005 & 0.012 & 11.134 & 9.745 & 8.816 & 10.114 & 10.470 & 9.096 & 9.948 & \textbf{0.130} & 0.156 & 0.291 & 0.141 \\
PFSP & ta008 & 0.011 & 10.301 & 9.706 & 9.624 & 10.101 & 9.773 & 9.402 & 9.839 & 0.135 & 0.151 & 0.290 & \textbf{0.133} \\
PFSP & ta013 & 0.022 & 22.766 & 20.120 & 19.688 & 21.923 & 21.348 & 21.357 & 23.793 & 0.215 & 0.230 & 0.381 & \textbf{0.200} \\
PFSP & ta018 & 0.022 & 20.839 & 21.915 & 18.532 & 21.302 & 19.753 & 20.889 & 18.767 & 0.208 & 0.232 & 0.353 & \textbf{0.204} \\
PFSP & ta021 & 0.045 & 48.317 & 47.494 & 42.107 & 48.533 & 48.922 & 49.204 & 44.547 & \textbf{0.369} & 0.402 & 0.521 & 0.384 \\
PFSP & ta025 & 0.047 & 50.547 & 44.339 & 43.591 & 43.709 & 45.344 & 48.325 & 49.076 & \textbf{0.378} & 0.387 & 0.492 & 0.380 \\
PFSP & ta038 & 0.042 & 46.883 & 44.621 & 40.681 & 43.742 & 40.036 & 43.401 & 44.608 & \textbf{0.267} & 0.299 & 0.461 & 0.274 \\
PFSP & ta039 & 0.046 & 46.747 & 48.209 & 45.814 & 45.625 & 43.179 & 49.648 & 43.738 & \textbf{0.262} & 0.292 & 0.429 & 0.269 \\
PFSP & ta041 & 0.083 & 96.011 & 93.193 & 85.334 & 94.707 & 94.174 & 97.375 & 90.536 & 0.478 & 0.503 & 0.657 & \textbf{0.440} \\
PFSP & ta049 & 0.085 & 101.031 & 93.844 & 88.639 & 99.260 & 93.994 & 83.248 & 103.776 & 0.491 & 0.488 & 0.672 & \textbf{0.444} \\
PFSP & ta059 & 0.182 & 204.231 & 208.579 & 184.028 & 224.158 & 206.053 & 205.553 & 204.084 & \textbf{1.037} & 1.054 & 1.273 & 1.167 \\
PFSP & ta060 & 0.189 & 217.622 & 185.284 & 215.390 & 223.717 & 201.794 & 186.349 & 224.066 & \textbf{1.033} & 1.065 & 1.275 & 1.115 \\
PFSP & ta064 & 0.169 & 156.168 & 145.219 & 149.630 & 164.002 & 173.886 & 152.962 & 153.102 & 0.525 & 0.537 & 0.717 & \textbf{0.483} \\
PFSP & ta069 & 0.144 & 157.044 & 146.140 & 158.429 & 160.079 & 153.738 & 158.015 & 160.673 & 0.476 & 0.499 & 0.706 & \textbf{0.470} \\
\end{longtable}
}}

% Resume the article layout after the wide benchmark tables.
\twocolumn

\subsection{Search settings}\label{sec:search-settings}

APEX evolutionary searches use Pareto survival.
XH uses mutation and crossover probabilities of 0.75 and 0.25, population
size 16 and an internal archive limit of 32. XT uses branching 4, depth 8
and exploration coefficient 1.4. XE uses population size 16, binary-tournament
parent selection, bandit discount 0.99, exploration coefficient 1.0 and
uniform-exploration share 0.1. When both operator kinds are applicable,
crossover is selected with probability 0.5. One operator is applied per
offspring; XH's variation probabilities do not govern XE. Applicability
filtering and duplicate screening are active in every XE configuration.

All searches receive a total of 1,000 complete-candidate evaluations;
two-phase combinations allocate 500 per phase, and XHTE allocates
333, 333 and 334. XG performs one construction. XE receives a single
validated incumbent from preceding phases. Seeds are 19, 42 and 73;
all methods use one worker within each of the eight concurrent processes,
executed in a fixed shuffled order.

The external methods use population size 100. SMS-EMOA produces batches
of 100 offspring in the evaluated configuration. MOEA/D uses 100 evenly
spaced reference directions, 20 neighbours and Tchebicheff decomposition
with its own update procedure. Duplicate handling follows the library
defaults for NSGA-II, SPEA2 and SMS-EMOA and is disabled for MOEA/D.
Crossover and mutation probabilities are 0.9 per mating and 1.0 per
offspring, respectively. Complete resolved configurations are retained
with the run records.

\subsection{Code and reproducibility}\label{sec:reproducibility}

The host was an Intel Core Ultra 9 275HX with 24 logical processors. Internal
wall time excludes package imports, raw-source parsing, final export and the
post-run audit; process startup latency is recorded separately.

The algorithm benchmark code and frozen reproduction archive are available at
\url{https://github.com/qunevo/apex/tree/main/benchmark}. The archived source
package and hashes identify the measured implementation and harness
independently of subsequent application changes.

Run records preserve source provenance, seeds, settings, evaluation counts,
timings and trade-off solutions. Tables and figures are regenerated from these
records, checking seed medians and aggregate arithmetic without new solver
runs. The independent audit checks operation coverage, machine eligibility,
durations, precedence, resource non-overlap, the PFSP permutation, objective
values and archive hypervolume. Environment details and source fingerprints
remain in the reproduction material.

\subsection{Synthetic workflow protocol and measured effort}\label{sec:workflow-protocol}

\paragraph{Factory and tasks.}
The six-job fixture in Table~\ref{tab:workflow-fixture} uses seconds from
an arbitrary planning origin, an eight-hour horizon and release time zero
for every operation. Each job's first operation precedes its second.
M1--M4 each have unit capacity and are initially available throughout the
horizon. Operations cannot be interrupted. J1-1 has a hard completion
deadline at 3,600~s. Only the second operation of each job has a soft due
date, initially with priority one. Each operation costs four synthetic units on M1 or M3 and one on M2
or M4, irrespective of duration. Setup, transport and material restrictions
are absent. Weighted operation tardiness sums positive completion-minus-due
differences multiplied by operation priority.

\begin{table}[htbp]
\centering\small
\caption{Entirely synthetic workflow fixture. Entries under M1--M4 are
alternative processing times in seconds, not consecutive visits to both
machines. Each job uses one machine per stage.}
\label{tab:workflow-fixture}
\begin{tabular}{@{}lrrrrr@{}}
\toprule
& \multicolumn{2}{c}{Stage 1} & \multicolumn{2}{c}{Stage 2} & \\
\cmidrule(lr){2-3}\cmidrule(lr){4-5}
Job & M1 & M2 & M3 & M4 & Stage-2 due time \\
\midrule
J1 & 1,200 & 1,500 & 900 & 1,200 & 4,200 \\
J2 & 900 & 1,200 & 1,500 & 900 & 4,800 \\
J3 & 1,800 & 1,200 & 1,200 & 1,500 & 5,400 \\
J4 & 1,500 & 900 & 1,800 & 1,200 & 6,000 \\
J5 & 1,200 & 1,800 & 900 & 1,500 & 6,600 \\
J6 & 900 & 1,500 & 1,200 & 900 & 7,200 \\
\bottomrule
\end{tabular}
\end{table}

\paragraph{Agent and checks.}
The sessions used GPT-6-astra at the \texttt{xhigh} reasoning setting,
the shared planning instructions, a generic reference for supported model
changes, and study-specific tool limits. Reference answers and verifier
code were not supplied to the agent. Sessions ran sequentially, in fixed
request order within each repetition, with fresh contexts and stores.
Each allowed 180~s, 32 tool calls and at most two XH searches of 64 candidate
evaluations, population eight, seed 42 and one worker. Each actual session
used one such search. Shell access, web access, subagents and code generation
were disabled. Preparing the baseline, scripted controls and final external
audit was outside session timing.

The independent checker did not import the native schedule validator.
It checked task coverage, eligible machines, durations, resource occupation,
precedence, windows, locks and recalculated metrics, as well as the requested
model changes and unchanged fields. It also checked scenario lineage,
schedule revision, baseline preservation, a successful native validation
call and, where requested, the comparison call and makespan arithmetic.
Eight scripted reference workflows and 18 deliberately invalid or no-op
controls checked the evaluation procedure before the measured sessions. The fixture,
case definitions, guidance, source, binary and verifier were frozen;
raw traces and stored artifacts were retained. The experiment resided in
an optional customization package and changed no scheduling-core code.

\paragraph{Measured effort.}
Table~\ref{tab:workflow-effort} reports interaction and search effort for
the 24 sessions in Table~\ref{tab:workflow-followup}. Internal XH time measures
only search, whereas session time includes the agent and tool round trips.
Input-token counts accumulate context across model calls and include cached
input; they are not billing estimates.

\begin{table}[htbp]
\centering\small\setlength{\tabcolsep}{5pt}
\caption{Measured effort for the 24 configuration and planning sessions.
Token usage was recorded for every session. Input counts include cached
context and repeated input across model calls.}
\label{tab:workflow-effort}
\begin{tabular}{@{}lr@{}}
\toprule
Measure & Value \\
\midrule
Sessions & 24 \\
Median session time (s) & 44.5 \\
Median tool calls & 7 \\
Median internal XH time (ms) & 16.7 \\
Median input tokens & 163,176.5 \\
Median output tokens & 770.5 \\
\bottomrule
\end{tabular}
}
\end{table}
\FloatBarrier

The study recorded 3,915,098 input tokens, including 3,255,296 cached
input tokens, and 17,957 output tokens across its 24 sessions, with
17.6 minutes of summed session time.
The preserved run records bind these measurements to source and
artifact hashes; the manuscript tables were produced by a read-only re-audit
without executing new agent or solver runs.
\FloatBarrier

\subsection{Native-extension implementation protocol}\label{sec:native-extension-protocol}

\paragraph{Requests and fixtures.}
The coding experiment has two separately evaluated requests: add the exact
squared-tardiness objective and add the hard exposure-balance rule defined
in Section~\ref{sec:native-extension-study}. The public examples each have
four one-operation jobs, two unit-capacity machines, a horizon of 100~s,
zero releases and no preemption. In the objective example, alternative
M1/M2 durations are $(4,6)$, $(1,2)$, $(3,5)$ and $(2,3)$; priorities are
$(1,3,2,1)$ and soft due times $(3,1,6,4)$. In the constraint example,
every operation takes 1~s on M1 or 2~s on M2 and contributes two exposure
points. Both machines are configured and $\Delta=0$ requires equal totals.
Makespan remains the original objective, becoming secondary only for the
objective request. Exposure is a synthetic allocation attribute, not a
validated physiological measurement.

Both requests also face two concealed variants. One has five renamed jobs
and two renamed machines, varied durations and exposures, a missing due
date and a zero priority. The other has four jobs with changed data,
a release at 3~s, a machine unavailable from 3 to 5~s, a precedence
relation and a machine commitment. Exact input records accompany the
frozen source. Additional probes isolate product-ready versus processing
completion, missing due dates, zero and fractional priority, invalid rule
configuration, missing or negative exposure, equality at the limit,
an unused configured resource and forced violations.

\paragraph{Coding and execution.}
The evaluation comprises seven sequential sessions with GPT-6-astra at
\texttt{xhigh}: three objective and four constraint attempts, each in a fresh
context. The agent received a precise requirement, the
existing extension skill and Rust interface declarations. It could only
submit a module to a tool that formatted and compiled it, then ran the
public example. Shell, web, subagents, concealed tests and reference code
were inaccessible. Each session allowed four such calls. Six sessions had a
300~s limit; an additional constraint attempt had a 900~s limit after one
session ended without a submission. It used the identical prompt, public
example, library and checker and received no previous implementation or
concealed-test result. It completed in 299.95~s and passed all 61 checks on
independent re-audit. The larger allowance does not establish a causal effect
on completion. Compiler or public-example failures could be repaired within
the session budget; there were no human edits to generated code.

An isolated host supplied the customization to the unchanged library.
Each primary XH run used 64 candidate evaluations, population eight, seed
42 and one worker, with existing due/slack/priority Qs as its initial policy.
Offline XT and XE runs checked compatibility at the same evaluation limit;
two-worker XH checked repeatability of assignments and score. These checks
are not another complete algorithm comparison. Compilation reused cached
dependencies in a release-profile host with optimisation level one;
its internal times are not directly comparable to the main benchmark.

\paragraph{Acceptance and controls.}
The deterministic Python checker does not import the Rust schedule
validator. It checks coverage, machine eligibility, durations, product
readiness, calendars, resource occupation, precedence, releases and
commitments, then independently calculates the requested criterion.
It checks that the new objective enters fitness while preserving the
existing objective, and that the hard rule rejects otherwise feasible
violating plans. Native validation, deliberate metric corruption and
independently corrupted schedule checks supplement these comparisons.
The complete suites contain 68 assertions for an objective implementation
and 61 for a constraint implementation.

Scripted references passed before coding. A linear-instead-of-squared
objective and a constraint with disabled final validation both compiled
but failed the checker. During preflight, final rejection alone did not
find the public balanced plan within the XH allowance; the conservative
prefix-filter requirement was added before the scored protocol was frozen.

\paragraph{Observed effort and repairs.}
Among sessions that submitted a final module, median time was 198.0~s for the
objective and 249.6~s for the constraint. Median accumulated compilation time
was 3.56~s and 1.20~s, respectively.
Each objective session repaired an initial Rust type error and a subsequent
public-example failure caused by requiring a stored custom metric before
the host had attached it. The supplied API excerpt omitted the nested phase
type and did not describe this callback ordering. The record retains
3 failed compilations and 3 failed public examples;
final acceptance therefore demonstrates completion with feedback, not
first-attempt correctness. The session without a submission has no compiler
call, final token-usage record or code-correctness observation. It is retained
in the seven-attempt record and excluded from submitted-module acceptance
counts and their timing medians.

The six sessions with final submissions reported 568,029 input tokens,
including 416,640 cached input tokens, and 43,649 output tokens.
Input counts accumulate context across model calls; cached input is already
included. These are token counts, not currency estimates. Session time
includes generation and tool round trips, while compilation and internal
search times describe separate components.

Submissions, tool responses, final modules, schedules and check outcomes are
retained locally with source and binary fingerprints. Core source and manifest
hashes match the pre-run freeze. The saved modules can be recompiled and
rechecked without another coding session.

\FloatBarrier

\balance
\paragraph{AI assistance.}

AI tools assisted manuscript preparation and the development of the Rust
implementation. The algorithms and experimental framework build on several
years of research and development. Responsibility for the methods, verification
and interpretation of the results remains with the authors.

\end{document}